\documentclass{article}

\usepackage[accepted]{icml2024}

\usepackage{hyperref}
\usepackage{url}
\usepackage{subcaption,graphicx}
\usepackage{caption}
\usepackage{tabularray}
\usepackage{rotating}
\usepackage{array}
\usepackage{multirow}
\usepackage{ragged2e}
\usepackage{booktabs}
\usepackage{arydshln}
\newcommand{\norm}[1]{\left\lVert#1\right\rVert}
\newcommand{\dotP}[2]{\langle#1,#2\rangle}
\usepackage{amsmath}
\usepackage{amssymb}
\usepackage{mathtools}
\usepackage{dsfont}
\usepackage{amsthm}
\usepackage{bm}
\usepackage[thinc]{esdiff}

\theoremstyle{plain}
\newtheorem{theorem}{Theorem}[section]

\theoremstyle{definition}

\newtheorem{insight}[theorem]{Insight}

\theoremstyle{remark}

\newtheorem*{theorem*}{Statement}

\usepackage{comment}

\icmltitlerunning{GATE: How to Keep Out Intrusive Neighbors}

\begin{document}

\twocolumn[
\icmltitle{GATE: How to Keep Out Intrusive Neighbors}



\icmlsetsymbol{equal}{*}

\begin{icmlauthorlist}
\icmlauthor{Nimrah Mustafa}{cispa}
\icmlauthor{Rebekka Burkholz}{cispa}
% \icmlauthor{Firstname3 Lastname3}{comp}
\end{icmlauthorlist}

\icmlaffiliation{cispa}{CISPA Helmholtz Center for Information Security, 66123 Saarbrücken, Germany}
%\icmlaffiliation{comp}{Company Name, Location, Country}
%\icmlaffiliation{sch}{School of ZZZ, Institute of WWW, Location, Country}

\icmlcorrespondingauthor{Nimrah Mustafa}{nimrah.mustafa@cispa.de}
%\icmlcorrespondingauthor{Firstname2 Lastname2}{first2.last2@www.uk}

\icmlkeywords{GATs, GNNs, Attention, Graph Attention Networks, GNNs architecture}

\vskip 0.3in
]



\printAffiliationsAndNotice{} % otherwise use the standard text.

\begin{abstract}
Graph Attention Networks (GATs) are designed to provide flexible neighborhood aggregation that assigns weights to neighbors according to their importance. 
In practice, however, GATs are often unable to switch off task-irrelevant neighborhood aggregation, as we show experimentally and analytically. 
To address this challenge, we propose GATE, a GAT extension that holds three major advantages: 
i) It alleviates over-smoothing by addressing its root cause of unnecessary neighborhood aggregation. 
ii) Similarly to perceptrons, it benefits from higher depth as it can still utilize additional layers for (non-)linear feature transformations in case of (nearly) switched-off neighborhood aggregation. 
iii) By down-weighting connections to unrelated neighbors, it often outperforms GATs on real-world heterophilic datasets. 
To further validate our claims, we construct a synthetic test bed to analyze a model's ability to utilize the appropriate amount of neighborhood aggregation, which could be of independent interest.
\end{abstract}

\section{Introduction}

Graph neural networks (GNNs) \citep{Gori2005NewModel} are a standard class of models for machine learning on graph-structured data that utilize node feature and graph structure information jointly to achieve strong empirical performance, particularly on node classification tasks. 
Input graphs to GNNs stem from various domains of real-world systems such as social \citep{Bian2020Rumor}, commercial \citep{Zhang2020Inductive}, academic \citep{hamaguchi2017knowledge}, economic \citep{monken2021intTrade}, biochemical\citep{Kearnes2016Molecular}, physical \citep{Shlomi2021physics}, and transport \citep{Wu2019Wavenet} networks that are diverse in their node feature and graph structure properties.

The message-passing mechanism of GNNs \citep{gcnKipf,xu2018Powerful} involves two key steps: a transformation of the node features, and the aggregation of these transformed features from a node's neighborhood to update the node's representation during training. 
While this has proven to be largely successful in certain cases, it generally introduces some problems for learning with GNNs, the most notorious of which is over-smoothing \citep{oversmoothing}. 
The enforced use of structural information in addition to node features may be detrimental to learning the node classification task, as shown by recent results where state-of-the-art GNNs perform the same as or worse than multi-layer perceptrons (MLPs) \citep{gomes2022when,yan2022sides,ma2023homophily}. 
One such task is where node labels can be easily determined by informative node features and require no contribution from the neighborhood. Here, standard neighborhood aggregation, as in most GNN architectures, would impair model performance, particularly with an increase in model depth.

A popular standard GNN architecture that, in principle, tries to resolve this problem is the Graph Attention Network (GAT) \citep{gatv1,gat}. By design, neighborhood aggregation in GATs is characterized by learnable coefficients intended to assign larger weights to more important neighboring nodes (including the node itself) in order to learn better node representations. 
Therefore, in the above example, GATs should ideally resort to assigning near-zero importance to neighbor nodes, effectively switching off neighborhood aggregation. 
However, we find that, counter-intuitively, GATs are unable to do this in practice and continue to aggregate the uninformative features in the neighborhood which impedes learning.% which impairs the performance of GAT, particularly with an increase in model depth.
%We construct a synthetic task (see Fig. \ref{examples-self-sufficient}) solely to evaluate the attention mechanism of GATs and its ability to focus on node features rather than neighborhood information or vice versa. 

One may ask why one would employ a GAT (or any GNN architecture) if an MLP suffices. 
%In practice, however, it is unknown a-priori whether graph and neighborhood information are useful for a learning task. Even if an MLP outperforms a specific GNN architecture, a different GNN model or intricate combinations of GNN layers and nonlinear feature transformations as performed by MLPs could define more adequate models. Ideally, neural network models can learn the appropriate degree of message passing. 
In practice, we do not know whether neighborhood aggregation (of raw features or features transformed by a perceptron or MLP), would be beneficial or not beforehand. 
This raises a pertinent question for the GNN research community: \textit{How much neighborhood aggregation is needed for a given task?}. Ideally, it is what we would want a model to learn. Otherwise, the right task-specific architecture would need to be identified by time and resource-intensive manual tuning. 

We address the challenge faced by GAT to effectively determine how well a node is represented by its own features compared to the features of neighboring nodes, i.e., distinguish between the relative importance of available node features and graph structure information for a given task. %how representative of a node are its own features and neighboring nodes' features, 

Firstly, we provide an intuitive explanation for the problem based on a conservation law of GAT gradient flow dynamics derived by \citet{balanceGATs}. %\todo{cite neurips paper}. 
Building on this insight, we present GATE, an extension of the GAT architecture that can switch neighborhood aggregation on and off as necessary. This allows our proposed architecture to gain the following advantages over GAT:
\begin{enumerate}
    \item It alleviates the notorious over-smoothing problem by addressing the root cause of unnecessarily repeated neighborhood aggregation.
    \item It allows the model to benefit from more meaningful representations obtained solely by deeper non-linear transformations, similarly to perceptrons, in layers with little to no neighborhood aggregation.% where neighborhood aggregation is is (nearly) switched off.
    \item It often outperforms GATs on real-world heterophilic datasets by weighing down unrelated neighbors.
    \item It offers interpretable learned self-attention coefficients, at the node level, that are indicative of the relative importance of feature and structure information in the locality of the node.
\end{enumerate}

In order to validate these claims, we construct a synthetic test bed of two opposite types of learning problems for node classification where label-relevant information is completely present only in a node's i) own features and ii) neighboring nodes' features (see Fig. \ref{examples}). 
GATE is able to adapt to both cases as necessary. 
On real-world datasets, GATE performs competitively on homophilic datasets and is substantially better than GAT on heterophilic datasets. %(see Table \ref{real-het-results}). 
Furthermore, up to our knowledge, it achieves a new state of the art on the relatively large OGB-arxiv dataset \cite{ogb} (i.e., $79.57\pm0.84$\% test accuracy). In summary, our contributions are as follows:
\begin{itemize}
    \item We identify and experimentally demonstrate a structural limitation of GAT, i.e., its inability to switch off neighborhood aggregation.
    \item We propose GATE, an extension of GAT, that overcomes this limitation and, in doing so, unlocks several benefits of the architecture.
    \item We update an existing conservation law relating the structure of gradients in GAT to GATE.
    \item We construct a synthetic test bed to validate our claims, which could be of independent interest to measure progress in developing adaptive neighborhood aggregation schemes. %given the active research along similar lines.
\end{itemize}

\begin{figure}[t]
\centering
\includegraphics[width=.45\textwidth]{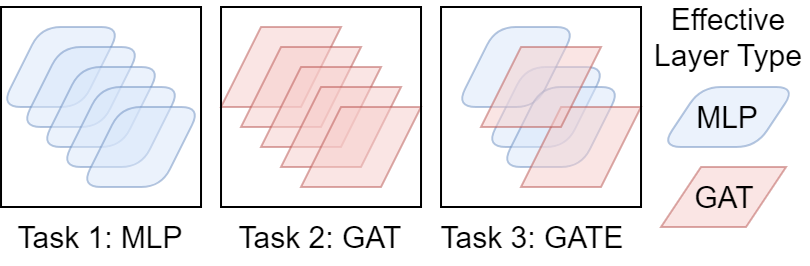}
  \caption{MLP only performs node feature transformations, whereas GAT also always aggregates over the neighborhood. With the ability to switch off neighborhood aggregation, GATE can learn to emulate MLP behavior and potentially interleave effective perceptron and standard layers in a flexible manner. This allows for more expressive power that we find to benefit real-world tasks (see Table \ref{real-het-results}).}% than predefined layers rigidly embedded in the architecture. }
  \label{layerTypes}
\end{figure}

 % The advantage of GATE is that it is able to effectively learn where to place GAT or perceptron layers or a mixture, as it fits the task at hand. Specifically, GATE emulates an MLP layer by only performing the non-linear transformation of the node's own features without then aggregating the transformed representations over the neighborhood. As we have shown, a GCN or GAT is unable to restrict the network to perform only the non-linear transformation step. Therefore, empowering the network to simulate both perceptron and standard GNN behavior allows for more expensive power instead of predefining the role. Effective perceptron and standard layers could potentially be interleaved by the model in a flexible manner, rather than being rigidly embedded in the architecture. We observe this, particularly for the heterophilic datasets in Figure 4,

%Input graphs with varying structural properties and feature information add to this difficulty. %describe the MLP-GNN-MLP-GNN block layers architecture

\section{Related Work}

To relieve GNNs from the drawbacks of unnecessarily repeated neighborhood aggregation in deeper models, initial techniques were inspired by classical deep learning of MLPs such as normalization \citep{graphnorm,pairnorm,groupnorm,nodenorm} and regularization \citep{dropgnn,dropedge,propreg,ladies}.

More recently, the need for deeper models and architectural changes to limit neighborhood aggregation as necessary has been recognized. Some approaches use linear combinations of initial features and current layer representation \citep{appnp}, 
add skip connections and identity mapping \citep{chen2020simple,depthBenefitsGNNs}, 
combine representations of all previous layers at the last layer \citep{jknets},
aggregate information from a node-wise defined range of $k$-hop neighbors\citep{dagnn}, and
limit the number of aggregation iterations based on node influence scores \citep{ndls}.
However, these architectures are not flexible enough to utilize additional network layers to simulate perceptron behavior, which, as we find, benefits heterophilic tasks. \cite{Ma2022homophily} discuss `good' and `bad' heterophily, which are also task-dependent. 

Other contemporary works for general GNNs propose the use of bilevel optimization to determine a node's strategic discrete action to a received message \citep{coopGNNs} and a variational inference framework for adaptive message passing \citep{AMP}. While these non-attentive architectures improve message passing in GNNs, we focus on identifying and explaining a structural limitation of self-attention in GAT, that continues to be used as a strong baseline architecture.

An orthogonal line of research uses graph structural learning \citep{yang2019top,
stretcu2019agree, franceschi2020learning} to amend the input graph structure such that neighborhood aggregation benefits the given task. 
Such approaches are difficult to scale, more susceptible to over-smoothing, and potentially destroy any inherent information in the original graph structure. 
On the contrary, a standard GNN architecture equipped to selectively perform neighborhood aggregation avoids these pitfalls. Self-supervision of the attention mechanism has also been proposed \citep{wang2019improving,superGAT}. Methods such as graph rewiring \citep{deac2022expander} to overcome problems such as over-squashing \citep{oversquashing} are complementary and may also be combined with GATE. %Additional supervision has also been proposed to improve the attention mechanism in GATs \citep{wang2019improving,superGAT}.% which also hints towards trainability issues that we discuss.

While we focus our insights on GAT, architectures based on GAT such as $\omega$GAT \citep{wGAT} also suffer from the same problem (see Fig. \ref{nbr-aggr-wGAT} in Appendix \ref{appendix:additional-results}). This further confirms that the universal problem with GAT has been correctly identified. In general, recent works direct effort to understand the current limitations of graph attention \citep{lee2023deep,fountoulakis2023graph}. 

 % \textcolor{blue}{Recently, there has been an increased interest in understanding and addressing the current limitations of graph attention \citep{lee2023deep,fountoulakis2023graph}. While we focus on the task of switching off irrelevant neighborhood aggregation, or `bad' heterophily, an effective attention mechanism should, ideally, also be able to distinguish between `good' and `bad' heterophily \citep{Ma2022homophily}, the presence of which is also task-dependent. This work serves as a stepping stone to enable more effective attention mechanisms for GNNs in the future}.

% and is more universal for multiple architectures. 

%%%An effective attention mechanism would also be able to distinguish between `good' and `bad' heterophily \citep{Ma2022homophily}, the presence of which is also task-dependent.

\section{Architecture}

\paragraph{Notation} 
Consider a graph $G=(V,E)$ with node set $\mathbb{V}$ and edge set $\mathbb{E} \subseteq \mathbb{V}\times \mathbb{V}$, where for a node $v \in \mathbb{V}$ the neighborhood is $\mathbb{N}(v)=\{u \vert (u,v) \in \mathbb{E}\}$ and input features are $\mathbf{h}_v^0$. 
A GNN layer updates each node's representation by aggregating over its neighbors' representation and combining it with its own features. 
The aggregation and combination steps can be performed together by introducing self-loops in $G$ such that, $\forall v \in \mathbb{V}$, $(v,v) \in \mathbb{E}$. We assume the presence of self-loops in $G$ unless specified otherwise.
In GATs, this aggregation is weighted by parameterized attention coefficients $\alpha_{uv}$, which indicate the importance of node $u$ for $v$.
A network is constructed by stacking $L$ layers, defined as follows, using a non-linear activation function $\phi$ that is homogeneous (i.e $\phi(x) = x\phi'(x)$) and consequently, $\phi(ax)=a\phi(x)$ for positive scalars $a$) such as ReLU $\phi(x) = \max\{x,0\}$ or LeakyReLU $\phi(x) = \max\{x,0\} + -\alpha \max\{-x,0\}$. %used is homogeneous (i.e $\phi(x) = x\phi'(x)$ and consequently, $\phi(ax)=a\phi(x)$ for positive scalars $a$), 

\paragraph{GAT} 
Given input representations $\mathbf{h}_v^{l-1}$ for $v\in \mathbb{V}$, a GAT \footnote{Throughout, we refer to GATv2 \citep{gat} as GAT.} layer $l \in [L]$ transforms those to:
\begin{align}
    \mathbf{h}_v^{l} &= \phi \left(\sum_{u\in \mathbb{N}(v)} \alpha_{uv}^l \cdot \textbf{W}_s^{l} \textbf{h}_u^{l-1} \right), \;\;\text{where} 
    \label{GATdef1} \\
      % \alpha_{uv}^l &= \frac{\text{exp}((a^l)^\top \cdot \text{LeakyReLU}(W_s^l h_u^{l-1} + W_t^l h_v^{l-1}))}{\sum_{u'\in \mathcal{N}(v)}\text{exp}( (a^l)^\top \cdot \text{LeakyReLU}(W_s^l h_{u'}^{l-1} + W_t^l h_v^{l-1}) ))}, \;\;\text{and}  
      % \label{GATdef2} \\
       \alpha_{uv}^l &= \frac{\text{exp}\left(e_{uv}^l\right)}{\sum_{u'\in \mathbb{N}(v)}\text{exp}\left( e_{u'v}^l\right)}, \;\;\text{and}  
      \label{GATdef2} \\
      e_{uv}^l &= \left(\textbf{a}^l\right)^\top \cdot \phi \left(\textbf{W}_s^l \textbf{h}_u^{l-1} + \textbf{W}_t^l \textbf{h}_v^{l-1}\right)
      \label{GATdef3}
    %e(h_u,h_v) &= (a^l)^\top \cdot \text{LeakyReLU}(W_s^l h_u^{l-1} + W_t^l h_v^{l-1}) 
   % \label{GATdef3}
\end{align}

The feature transformation weights $\textbf{W}_{s}$ and $\textbf{W}_{t}$ for source and target nodes, respectively, may also be shared such that $\textbf{W}_{s}=\textbf{W}_{t}$. 
We denote this variant of GAT by GAT$_{S}$.

\paragraph{GATE}
In addition, we propose GATE, a GAT variant that flexibly weights the importance of node features and neighborhood features.
A GATE layer is also defined by Eq. (\ref{GATdef1}) and (\ref{GATdef2}) but modifies $e_{uv}$  in Eq. (\ref{GATdef3}) to Eq. (\ref{GATEdef}). 
%Given that $q_{uv}=1$ if $u=v$ and $q_{uv}=0$ if $u\neq v$, 
\begin{equation}
      e_{uv}^l = \left(\mathds{1}_{u\neq v}\textbf{a}_s^l + \mathds{1}_{u=v} \textbf{a}_t^l \right) ^\top \cdot \phi \left(\textbf{U}^l \textbf{h}_u^{l-1} + \textbf{V}^l \textbf{h}_v^{l-1}\right). \label{GATEdef} 
      %\;\;\text{where} \\       &= q_{uv}    
\end{equation}

% \begin{align}
%       e_{uv,u\neq v}^l &= \left(\textbf{a}_s^l\right)^\top \cdot \phi \left(\textbf{U}^l \textbf{h}_u^{l-1} + \textbf{V}^l \textbf{h}_v^{l-1}\right)
%       \label{GATEdef1}\\
%       e_{uv,u=v}^l &= 
%       \left(\textbf{a}_t^l\right)^\top \cdot \phi \left(\textbf{U}^l \textbf{h}_u^{l-1} + \textbf{V}^l \textbf{h}_v^{l-1}\right) = e_{vv}^l
%       \label{GATEdef2}  
%     %e(h_u,h_v) &= (a^l)^\top \cdot \text{LeakyReLU}(W_s^l h_u^{l-1} + W_t^l h_v^{l-1}) 
%    % \label{GATdef3}
% \end{align}

We denote $e_{uv}$ in Eq. (\ref{GATdef3}) and (\ref{GATEdef}) as $e_{vv}^l$ if $u=v$. For GATE, $\mathbf{W}_s^l$ in Eq. (\ref{GATdef1}) is denoted as $\mathbf{W}^l$. A weight-sharing variant of GATE, GATE$_S$, is characterized by all feature transformation parameters being shared in a layer (i.e. $\mathbf{W}^l=\mathbf{U}^l=\mathbf{V}^l$). Note that, then, for a $d$-dimensional layer, GATE adds only $d$ more parameters to GAT.% and thus has the same computational complexity as GAT.% and only introduces $d$ additional parameters.
 %Yet, this leads to a significant impact on generalization performance as we demonstrate next.

% \todo{update this paragraph depending on any added results}
% For discussion and ablation purposes, we consider two variants of GATE with weight-sharing. 
% Firstly, GATE$_S$ with all feature transformation parameters in Eq. (\ref{GATdef1}) and (\ref{GATEdef}) are shared, i.e. $\mathbf{W}=\mathbf{U}=\mathbf{V}$. 
% Secondly, GATE$_{SS}$ where only source node transformation  is shared, i.e. $\mathbf{W} = \mathbf{U} \neq \mathbf{V}$.

We next present theoretical insights into the reasoning behind the inability of GATs to switch off neighborhood aggregation, which is rooted in norm constraints imposed by the inherent conservation law for GATs. The gradients of GATE fulfill an updated conservation law  (Thoerem \ref{strucOfGradsGATE}) that enables switching off neighborhood aggregation in a parameter regime with well-trainable attention.

\section{Theoretical Insights}% into neighborhood aggregation}
%\citet{balanceGATs} have recently derived a conservation law that relates the parameters of a GAT and their gradients in the following way.
For simplicity, we limit our discussion here to GATs with weight sharing. We derive similar arguments for GATs without weight sharing in Appendix \ref{insightDer:GAT}.
The following conservation law was recently derived for GATs to explain trainability issues of standard initialization schemes.
Even with improved initializations, we argue that this law limits the effective expressiveness of GATs and their ability to switch off neighborhood aggregation when necessary.% hinders them from switching off neighborhood aggregation when necessary.

\begin{theorem}[Thm. 2.2 by \citet{balanceGATs}]
The parameters $\theta$ %(i.e., $\mathbf{W}^l$ for feature transformation and $\mathbf{a}^l$ for self-attention) 
of a layer $l\in[L-1]$ in a GAT network and their gradients $\nabla_{\theta} \mathcal{L}$ w.r.t. loss $\mathcal{L}$ fulfill:
\begin{equation}
\dotP{\mathbf{W}^l_{[i,:]}}{\nabla_{\mathbf{W}^l_{[i,:]}}} =\dotP{\mathbf{W}^{l+1}_{[:,i]}}{\nabla_{\mathbf{W}^{l+1}_{[:,i]}}} + \dotP{\mathbf{a}^l_{[i]}}{\nabla_{\mathbf{a}^l_{[i]}} }.
\label{eqStructureOfGrads}
\end{equation}
\label{theoremStructureOfGrads}
\end{theorem}
Intuitively, this equality limits the budget for the relative change of parameters and imposes indirectly a norm constraint on the parameters. 
Under gradient flow that assumes infinitesimally small learning rates, this law implies that the relationship $\norm{\mathbf{W}^l[i,:]}^2 - \norm{\mathbf{a}^l[i]}^2 -\norm{\mathbf{W}^{l+1}[:i]}^2 = c$ stays constant during training, where $c$ is defined by the initial norms. 
Other gradient-based optimizers fulfill this norm balance also approximately. 
Note that the norms $\norm{\mathbf{W}^l[i,:]}$ generally do not assume arbitrary values but are determined by the required scale of the output.
Deeper models are especially less flexible in varying these norms as deviations could lead to exploding or diminishing outputs and/or gradients.
In consequence, the norms of the attention parameters are also bounded.
Furthermore, a parameter becomes harder to change during training when its magnitude increases.
This can be seen by transforming the law with respect to the relative change of a parameter defined as $\Delta \theta = \nabla_{\theta}\mathcal{L}/\theta$ for $\theta \neq 0$ or $\Delta \theta = 0$ for $\theta=0$.% as follows.% and applying it to multiple layers.
\begin{align} 
\sum^{n_{l+1}}_{j=1} {\mathbf{W}^{l}_{ij}}^2 \Delta \mathbf{W}^{l}_{ij}  = 
 \sum^{n_{l+2}}_{k=1} {\mathbf{W}^{l+1}_{ki}}^2 \Delta \mathbf{W}^{l+1}_{ki} + {\mathbf{a}^{l}_{i}}^2 \Delta \mathbf{a}^{l}_{i}.
\label{lastLayerParamChange}
\end{align}
The higher the magnitude of an attention parameter $(\mathbf{a}^{l}_{i})^2$,  the smaller will be the relative change $\Delta \mathbf{a}^{l}_{i}$ and vice versa. %Otherwise, the higher magnitude of $\norm{\mathbf{a}^l}$ would be compensated for by large $\norm{\mathbf{W}^l}$ and relatively small $\norm{\mathbf{W}^{l+1}}$ %inducing larger $\nabla \mathbf{W}^{l+1}$ and smaller $\nabla \mathbf{W}^{l}$. 
This restricts the attention mechanism in the network to a less-trainable regime without converging to a meaningful model. We next explain why large $\norm{\mathbf{a}^l}$ are required to switch off neighborhood aggregation in a layer.   %as explained by the next insight.
%\C{Furthermore, attention layers that are closer to the input are more norm constrained.
%This is why we observe in experiments that only later GAT layers in deep models can reduce the amount of neighborhood aggregation, as can also be explained by the next insight.}
%     \label{eqNormPreservation} 

% Gradient 
% \begin{equation}
%      \norm{W^l[i,:]}^2 - \norm{a^l[i]}^2 -\norm{W^{l+1}[:i]}^2 = c,
%     \label{eqNormPreservation}
% \end{equation}
\begin{insight}[Effective expressiveness of GATs]\label{insight:GAT}
GATs are challenged to switch off neighborhood aggregation during training, as this requires the model to enter a less trainable regime with large attention parameters $\norm{\mathbf{a}}^2 >> 1$.
\end{insight}
An intuitive derivation of this insight is presented in the appendix. 
Here, we outline the main argument based on the observation that to make the contribution of neighbor $j$ insignificant relative to node $i$, we require $\alpha_{ij}/\alpha_{ii} << 1$. 
We use relative $\alpha_{ij}/\alpha_{ii}$ instead of $\alpha_{ij}$ and $\alpha_{ii}$ to cancel out normalization constants and simplify the analysis.

Our key observation is that, given an insignificant link $(i,j)$, its relative contribution to its two neighborhoods $\alpha_{ij}/\alpha_{ii} << 1$ and $\alpha_{ji}/\alpha_{jj} << 1$ are affected in opposite ways by a feature $f$ of the attention parameters $\mathbf{a}$, i.e. if $\mathbf{a}[f]$ contributes to reducing $\alpha_{ij}/\alpha_{ii}$, it automatically increases $\alpha_{ji}/\alpha_{jj}$. 
However, we require multiple features that contribute to reducing only $\alpha_{ij}$ without strengthening $\alpha_{ji}$ that may only be possible in a high dimensional space requiring large norms of $\mathbf{a}$.
Yet, the norms $\norm{\mathbf{a}}^2$ are constrained by the parameter initialization and cannot increase arbitrarily due to the derived conservation law. 
Note that to switch off all neighborhood aggregation, we require $\alpha_{ij}/\alpha_{ii} << 1$, $\forall$ $j\in \mathbb{N}(i)$, further complicating the task.

% The main argument rests on the observation that to make the contribution of neighbor $j$ insiginificant relative to that of the node itself, we require $\alpha_{ij}/\alpha_{ii} << 1$. 
% The relative contribution of a link to its two neighborhoods $\alpha_{ij}/\alpha_{ii} << 1$ and $\alpha_{ji}/\alpha_{jj} << 1$ can only be small simultaneously for large norms $\norm{\mathbf{a}}^2 >>1$ with multiple features that contribute to $\alpha_{ij}$.
% Yet, the norms $\norm{\mathbf{a}}^2$ are constrained by the parameter initialization and cannot increase arbitrarily due to the derived conservation law.

To address this challenge, we modify the GAT architecture by GATE that learns separate attention parameters for the node and the neighborhood contribution.
As its conservation law indicates, it can switch off neighborhood aggregation in the well-trainable parameter regime. 
\begin{theorem}[Structure of GATE gradients]\label{theoremStructureGATE}
The parameters and gradients of a GATE network w.r.t. to loss $\mathcal{L}$ for layer $l \in [L-1]$ are conserved according to the following laws.
Given $\Theta(\theta)=\dotP{\theta}{\nabla_{\theta}\mathcal{L}}$, it holds that:  
\begin{equation}
    \Theta ( \mathbf{W}^l_{[i,:]} ) - \Theta( {\mathbf{a}^{l+1}_s}_{[i]} ) - \Theta( {\mathbf{a}^{l+1}_t}_{[i]} ) = \Theta( \mathbf{W}^{l+1}_{[:,i]} ).
    \label{eqStructureOfGradsGATE}
\end{equation}
and, if additional independent matrices $\mathbf{U}^l$ and $\mathbf{V}^l$ are trainable, it also holds  that:
\begin{equation}
\Theta({\mathbf{a}^l_s}_{[i]}) + \Theta ({\mathbf{a}^l_t}_{[i]}) = \Theta({\mathbf{U}^{l}}_{[i,:]}) + \Theta ({\mathbf{V}^{l}}_{[i,:]}). 
\label{eqStructureOfGradsGATE2}
\end{equation}
\label{strucOfGradsGATE}
\end{theorem}
The proof is provided in the appendix.
We utilize this theorem for two purposes.
Firstly, it induces an initialization that enables at least the initial trainability of the network.
Similarly to GAT \citep{balanceGATs}, we initialize all $\mathbf{a}$ parameters with zero and matrices $\mathbf{W}$ with random orthogonal looks-linear structure in GATE.
This also ensures that we have no initial inductive bias or preference for specific neighbor or node features. As an ablation, we also verify that the initialization of the attention parameters in GAT with zero alone can not switch off neighborhood aggregation in GAT (see Fig. \ref{gat-zero-attn} in Appendix \ref{appendix:additional-results}).

Secondly, the conservation law leads to the insight that a GATE network is more easily capable of switching off neighborhood aggregation or node feature contributions in comparison with GAT.

\begin{insight}[GATE is able to switch off neighborhood aggregation.]\label{insight:GATE}
GATE can flexibly switch off neighborhood aggregation or node features in the well-trainable regime of the attention parameters.
\end{insight}
This insight follows immediately from the related conservation law for GATE that shows that $\mathbf{a}_t$ and $\mathbf{a}_s$ can interchange the available budget for relative change among each other. Furthermore, the contribution of neighbors and the nodes are controlled separately. We  show how the respective switch-off can be achieved with relatively small attention parameter norms that correspond to the well-trainable regime in Appendix \ref{insightDer:GATE}.
To verify these insights in experiments, we next design synthetic data generators that can test the ability of GNNs to take structural infromation into account in a task-appropriate manner. 

%telescoping result by applying eq(6) to multiple layers, for a layer l which explains why all layers do attempt reduce the amount of neighborhood aggregation, but due to the constraints of the law, the later layers are more successful in doing so than the earlier ones. (and refer to figure in appendix). 
%$$ \sum_{j=1}^{n_{1}} \sum_{m=1}^{n_{0}} {W_{jm}^{(1)}}^2 \Delta W_{jm}^{(1)}   = \sum_{i=1}^{n_{L-1}} \sum_{k=1}^{n_L} {W_{ki}^{(L)}}^2 \Delta W_{ki}^{(L)} + \sum_{l=1}^{L-1} \sum_{o=1}^{n_{l}} {a_{o}^{(l)}}^2 \Delta a_{o}^{(l)} $$
%how the greater the depth, the greater number of attention parameters (summed over all layers) over which the difference in norms of Ws of first and last layer is distributed, hence the room for each attention parameter to change reduces as the number of parameters increases with depth. Does this also explain why GATE has a tendency to switch off later layers (t owards the end) instead of the earlier ones (at the start)? (as attention parameters in later layers have more room to change than earlier layers- only observed in models for real data (10 or more layers) but not on synthetic data (no 10-layer model run in this case).

\begin{figure}[t]
\centering
\begin{subfigure}{.5\textwidth}
  \centering
\includegraphics[width=.95\linewidth]{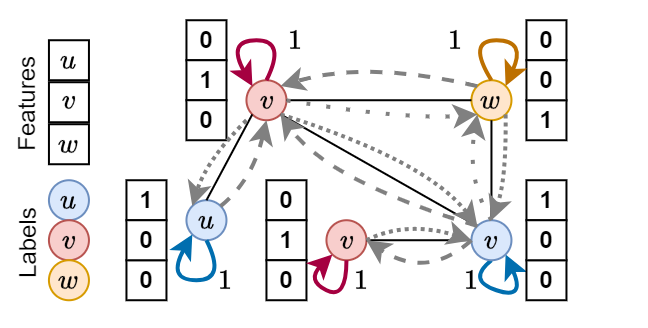}
\caption{No neighborhood contribution required.} 
  \label{examples-self-sufficient}
\end{subfigure}\\
\begin{subfigure}{.5\textwidth}
  \centering
\includegraphics[width=.95\linewidth]{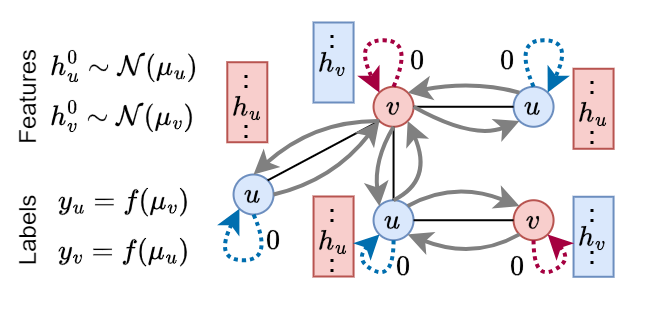}
\caption{Only neighborhood contribution required.} 
\label{examples-neighbor-dependent}
\end{subfigure}
  \caption{Examples of synthetic input graphs constructed for learning tasks that are (a) self-sufficient and can be better solved by switching off neighborhood aggregation, i.e. $\alpha_{vv}=1$ and (b) neighbor-dependent that benefit from ignoring the node's own features, i.e. $\alpha_{vv}=0$. In both cases, $\forall$ $v\in \mathbb{V}$, $\sum_{u\in\mathbb{N}(v),u\neq v} \alpha_{uv} + \alpha_{vv}=1$. These represent opposite ends of the spectrum whereas real-world tasks often lie in between and require $\alpha_{ii}\in [0,1]$. GATE's attention mechanism is more flexible than GAT's in learning the level of neighborhood aggregation required for a task. }
  \label{examples}
\end{figure}

\section{Experiments}\label{sec:experiments}

We validate the ability of GATE to perform the appropriate amount of neighborhood aggregation, as relevant for the given task and input graph, on both synthetic and real-world graphs. 
In order to gauge the amount of neighborhood aggregation, we study the distribution of $\alpha_{vv}$ values (over the nodes) at various epochs during training and layers in the network.
This serves as a fair proxy since $\forall$ $v \in \mathbb{V}$, $\alpha_{vv} = 1 - \sum_{u \in \mathbb{N}(v), u\neq v} \alpha_{uv}$.
Thus, $\alpha_{vv}=1$ implies no neighborhood aggregation (i.e. only $\mathbf{h}_v$ is used) whereas $\alpha_{vv}=0$ implies only neighborhood aggregation (i.e. $\mathbf{h}_v$ is not used). Figure \ref{examples} shows an exemplary construction of both these cases. We defer a discussion of the experimental setup to Appendix \ref{expSettings}.

% \paragraph{Experimental settings} 
% We vary the depth of GAT and GATE networks in our experiments, but keep the hidden layer width fixed to 64 in all cases. For GAT$_S$ and GAT networks, we substitute $\phi$ in Eq. (\ref{GATdef3}) with LeakyReLU as defined in the standard architecture. For GATE, we substitute $\phi$ in Eq. (\ref{GATEdef}) with ReLU in order to be able to interpret the sign of $\mathbf{a_s}$ and $\mathbf{a_t}$ parameters as contributing positively or negatively to neighborhood aggregation. We defer the remaining details of the experimental setup to Appendix \ref{expSettings}. All networks are trained using Adam.

\subsection{Synthetic Test Bed}

We construct the synthetic test bed as a node classification task for two types of problems: {\it self-sufficient} learning and {\it neighbor-dependent} learning. 
In the self-sufficient learning problem, complete label-relevant information is present in a node's own features. 
On the contrary, in the neighbor-dependent learning problem, label-relevant information is present in the node features of the $k$-hop neighbors. 
We discuss both cases in detail, beginning with the simpler one.% self-sufficient case. 

 \begin{table}[t]
\caption{Self-sufficient learning: $S, C$ and $L$ denote graph structure, number of label classes, and number of network layers, respectively. Original (Orig.) and randomized (Rand.) labels are used for the Cora structure. In all cases, $100\%$ train accuracy is achieved except in ones marked with $^\ast$ and GATE eventually achieves $100\%$ test accuracy a few epochs later except in one marked with $^\ddag$. GAT$_S$ and GAT models marked with $^\dag$ also eventually achieve $100\%$ test accuracy. GATE$_S$ behaves similarly to GATE and achieves $100\%$ train and test accuracy.}.%and also achieves $100\%$ train and test accuracy.}% \todo{add that for exps GATE achieves 100\% test accuracy a few (compute and add average) epochs after the epoch of min loss.}}
\label{table:self-suff-task}
\centering
\begin{tblr}{
  width = \linewidth,
  colspec = {Q[15]Q[15]Q[15]Q[155]Q[155]Q[155]},
  row{even} = {c},
  row{5} = {c},
  row{7} = {c},
  row{11} = {c},
  row{13} = {c},
  cell{1}{1} = {r=2}{},
  cell{1}{2} = {r=2}{c},
  cell{1}{3} = {r=2}{c},
  cell{1}{4} = {c=3}{0.732\linewidth,c},
  cell{3}{1} = {r=6}{},
  cell{3}{2} = {r=3}{c},
  cell{3}{3} = {c},
  cell{3}{4} = {c},
  cell{3}{5} = {c},
  cell{3}{6} = {c},
  cell{6}{2} = {r=3}{},
  cell{9}{1} = {r=6}{},
  cell{9}{2} = {r=3}{c},
  cell{9}{3} = {c},
  cell{9}{4} = {c},
  cell{9}{5} = {c},
  cell{9}{6} = {c},
  cell{12}{2} = {r=3}{},
  hline{1,15} = {-}{0.08em},
  hline{2} = {4-6}{0.03em},
  hline{3,9} = {-}{0.05em},
  hline{6,12} = {2-6}{0.03em},
}
S                                        & C                                 & L & Test Acc.@Epoch of Min. Train Loss &                    &                      \\
                                         &                                   &   & GAT$_S$                              & GAT                & GATE                 \\
\begin{sideways}Cora\end{sideways}       & \begin{sideways}O,$7$\end{sideways} & 1 & \textbf{99.1@215}$^\dag$                  & 97.7@166$^\dag$       & 99.0@127           \\
                                         &                                   & 2 & 93.4@218                           & 94.5@158         & \textbf{99.6@35}   \\
                                         &                                   & 5 & 85.9@92                            & 85.5@72          & \textbf{98.4@36}$^\ddag$   \\
                                         & \begin{sideways}R,$7$\end{sideways} & 1 & 99.4@263$^\dag$                            & 99.8@268$^\dag$          & \textbf{100@104}   \\
                                         &                                   & 2 & 61.7@2088$^\ast$                         & 52.8@341$^\ast$        & \textbf{99.9@36}   \\
                                         &                                   & 5 & 35.1@609                           & 32.1@1299        & \textbf{99.9@23}   \\
\begin{sideways}ER$(p=0.01)$\end{sideways} & \begin{sideways}R,$2$\end{sideways} & 1 & 100@341$^\dag$                             & \textbf{100@182}$^\dag$  & 100@1313           \\
                                         &                                   & 2 & 99.2@100$^\dag$                            & 99.2@119$^\dag$          & \textbf{99.6@79}   \\
                                         &                                   & 5 & 64.0@7778$^\ast$                         & 99.6@239         & \textbf{100@45}    \\
                                         & \begin{sideways}R,$8$\end{sideways} & 1 & 88.8@9578$^\ast$                         & 98.4@3290        & \textbf{99.2@1755} \\
                                         &                                   & 2 & 90.4@2459$^\ast$                         & 94.8@2237        & \textbf{99.6@44}   \\
                                         &                                   & 5 & 23.6@8152                          & 26.0@8121        & \textbf{100@28}    
\end{tblr}
\end{table}

\begin{figure} [t]
\centering
\includegraphics[width=\linewidth]{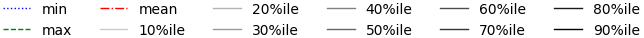}\\
\begin{subfigure}{.25\textwidth}
\begin{subfigure}{.33\textwidth}
  \centering
\frame{\includegraphics[width=.97\linewidth]{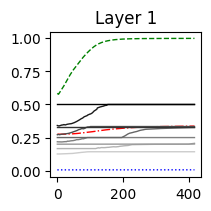}}
\end{subfigure}%
\begin{subfigure}{.66\textwidth}
  \centering  
\frame{\includegraphics[width=.95\linewidth]{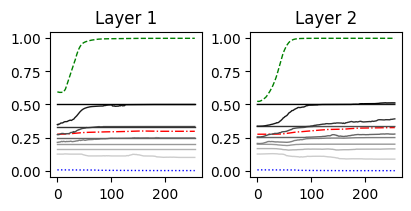}}
\end{subfigure}
\caption{GAT with original labels}
\end{subfigure}%
\begin{subfigure}{.25\textwidth}
\begin{subfigure}{.33\textwidth}
  \centering
\frame{\includegraphics[width=.97\linewidth]{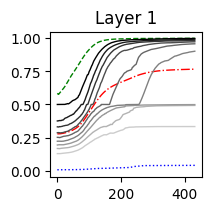}}
\end{subfigure}%
\begin{subfigure}{.66\textwidth}
  \centering  
\frame{\includegraphics[width=.95\linewidth]{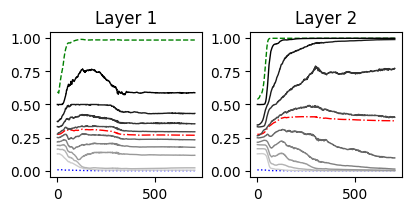}}
\end{subfigure}
\caption{GAT with random labels}
\end{subfigure}\\
\begin{subfigure}{.25\textwidth}
\begin{subfigure}{.33\textwidth}
  \centering
\frame{\includegraphics[width=.97\linewidth]{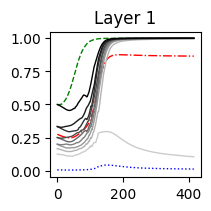}}
\end{subfigure}%
\begin{subfigure}{.66\textwidth}
  \centering  
\frame{\includegraphics[width=.95\linewidth]{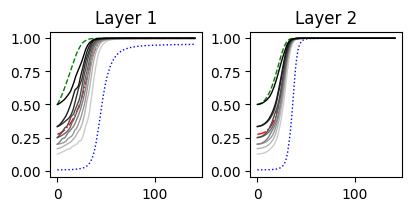}}
\end{subfigure}
\caption{GATE with original labels}
\end{subfigure}%
\begin{subfigure}{.25\textwidth}
\begin{subfigure}{.33\textwidth}
  \centering
\frame{\includegraphics[width=.97\linewidth]{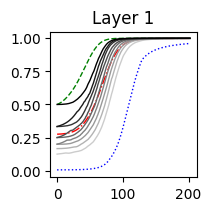}}
\end{subfigure}%
\begin{subfigure}{.66\textwidth}
  \centering  
\frame{\includegraphics[width=.95\linewidth]{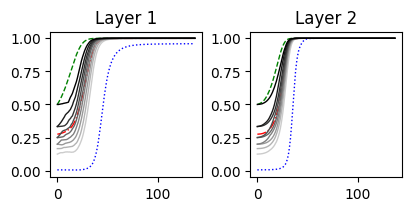}}
\end{subfigure}
\caption{GATE with random labels}
\end{subfigure}    
  \caption{ Distribution of $\alpha_{vv}$ against training epoch for self-sufficient learning problem using Cora structure and input node features as the one-hot encoding of labels for $1$ and $2$ layer models. Due to space limitation, we defer the plots of $5$ layer networks to Fig. \ref{alphaDist-cora-oneHotFeats-5-layer} in Appendix \ref{appendix:additional-results}.}
  \label{alphaDist-cora-oneHotFeats-1-and-2-layer}
\end{figure}

\begin{figure*}[t]
\centering
\begin{subfigure}{.33\textwidth}
  \centering
  \frame{
\includegraphics[width=0.5\linewidth]{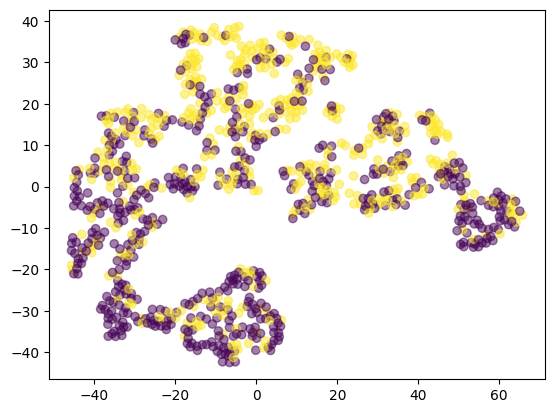}%
\includegraphics[width=0.5\linewidth]{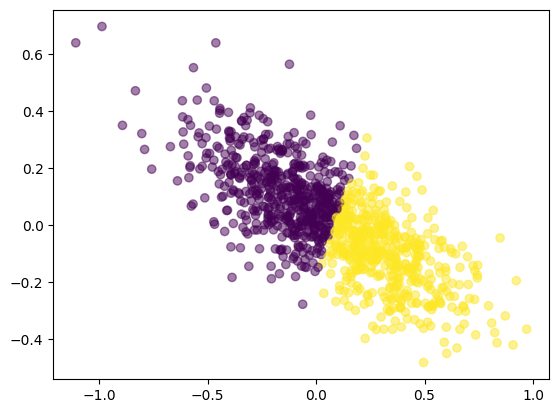}}
\caption{$k=1$} 
\label{}
\end{subfigure}%
\begin{subfigure}{.33\textwidth}
  \centering
\frame{\includegraphics[width=0.5\linewidth]{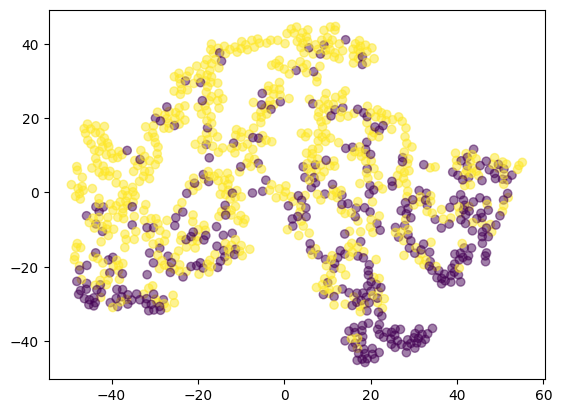}%
\includegraphics[width=0.5\linewidth]{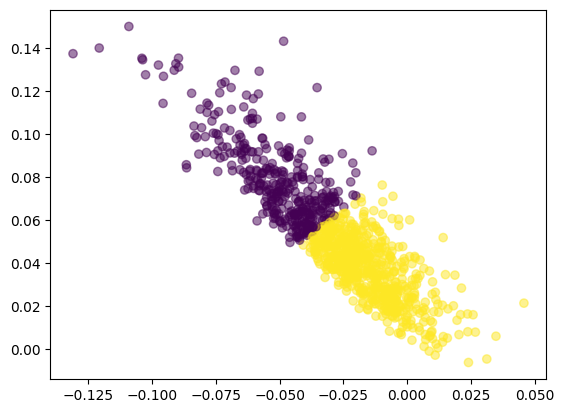}}
\caption{$k=2$} 
\label{}
\end{subfigure}%
\begin{subfigure}{.33\textwidth}
  \centering
\frame{\includegraphics[width=0.5\linewidth]{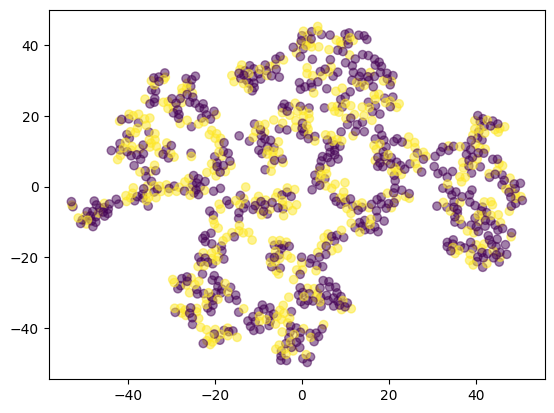}%
\includegraphics[width=0.5\linewidth]{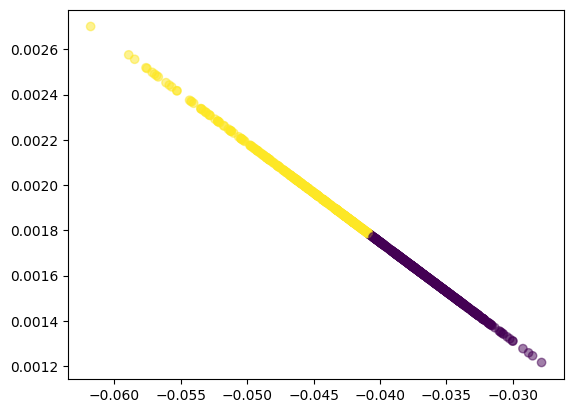}}
\caption{$k=3$} 
\label{}
\end{subfigure}
  \caption{(a)-(c): Distribution of node labels of a synthetic dataset, with neighbor-dependent node labels, based on nodes' own random features (left) and neighbors' features aggregated $k$ times (right).}
  \label{randSynData}
\end{figure*}

\begin{figure*} [t]
\centering
\includegraphics[width=.75\linewidth]{Figures/alphaDistribution/alpha_ii_distribution_legend.png}
\begin{subfigure}{.25\textwidth}
  \centering
\frame{\includegraphics[width=.99\linewidth]{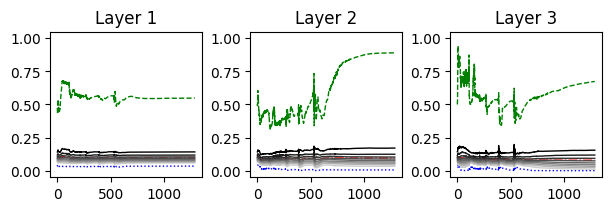}}
\end{subfigure}%
\begin{subfigure}{.33\textwidth}
  \centering  
\frame{\includegraphics[width=.98\linewidth]{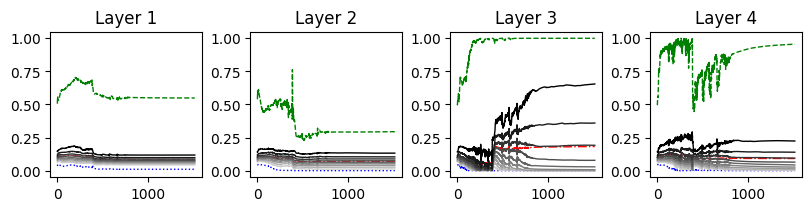}}
\end{subfigure}%
\begin{subfigure}{.42\textwidth}
  \centering  
\frame{\includegraphics[width=.97\linewidth]{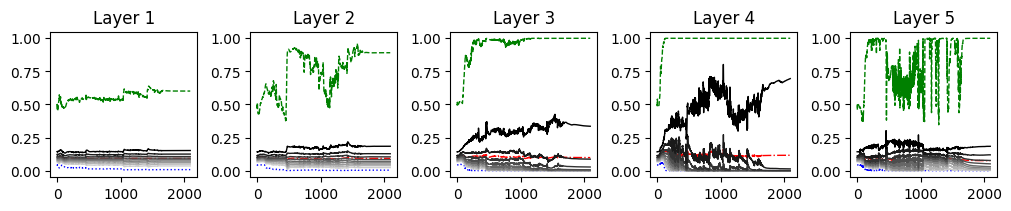}}
\end{subfigure}\\
\begin{subfigure}{.25\textwidth}
  \centering
\frame{\includegraphics[width=.99\linewidth]{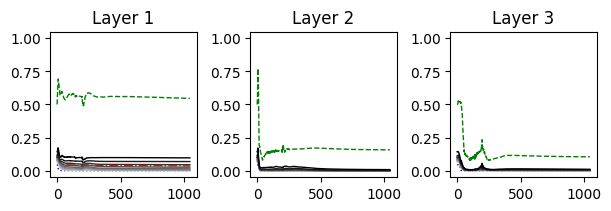}}
\end{subfigure}%
\begin{subfigure}{.33\textwidth}
  \centering  
\frame{\includegraphics[width=.98\linewidth]{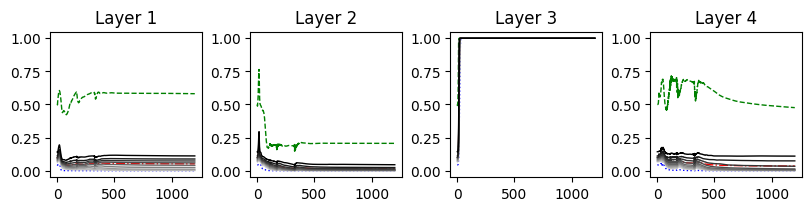}}
\end{subfigure}%
\begin{subfigure}{.42\textwidth}
  \centering  
\frame{\includegraphics[width=.97\linewidth]{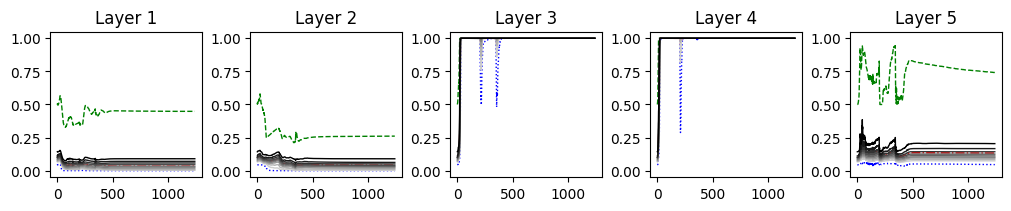}}
\end{subfigure}
  \caption{Distribution of $\alpha_{vv}$ against training epoch for the neighbor-dependent learning problem with $k=3$. Rows: GAT (top) and GATE (bottom) architecture. Columns (left to right): $3$, $4$, and $5$ layer models. While GAT is unable to switch off neighborhood aggregation in any layer, only $3$ layers of the $4$ and $5$ layer models perform neighborhood aggregation.}
  \label{alpha-dist-randSynData-k3}
\end{figure*}

\paragraph{Learning self-sufficient node labels} 
In order to model this task exactly, we generate an Erdős–Rényi $ ($ER) graph structure $G$ with $N=1000$ nodes and edge probability $p=0.01$. 
Node labels $y_v$ are assigned uniformly at random from $C=[2,8]$ classes. 
Input node features ${\mathbf h}_v^0$ are generated as one-hot encoded node labels in both cases, i.e., ${\mathbf h}_v^0 = \mathbf{1}_{y_v}$. Nodes are divided randomly into train/validation/test split with a $2:1:1$ ratio.

We also use a real-world graph structure of the Cora dataset. 
Two cases using this graph structure are tested: i) using the original node labels consisting of $7$ classes, and ii) randomized labels of $7$ classes. 
Input node features are generated as one-hot encoding of node labels in both cases. The standard splits of Cora are used.

As evident in Table \ref{table:self-suff-task}, GAT is unable to perfectly learn this task whereas GATE easily achieves $100\%$ train and test accuracy, and often in fewer training epochs. 

In line with the homophilic nature of Cora, GAT achieves reasonably good accuracy when the original labels of the Cora graph structure are used as neighborhood aggregation is relatively less detrimental, particularly in a single-layer model. 
Nevertheless, in the same case, GATE generalizes better than GAT with an increase in model depth. This indicates that over-smoothing, a major cause of performance degradation with model depth in GNNs, is also alleviated due to reduced neighborhood aggregations (see Fig. \ref{alphaDist-cora-oneHotFeats-1-and-2-layer}). 

On the contrary, random labels pose a real challenge to GAT. 
Since the neighborhood features are fully uninformative about a node's label in the randomized case, aggregation over such a neighborhood distorts the fully informative features of the node itself. 
This impedes the GAT network from learning the task, as it is unable to effectively switch off aggregation (see Fig. \ref{alphaDist-cora-oneHotFeats-1-and-2-layer}), whereas GATE is able to adapt to the required level of neighborhood aggregation (i.e. none, in this case).
Interestingly, note that a single layer GAT in the case of random labels can almost, though not completely, switch off neighborhood aggregation (see Fig. \ref{alphaDist-cora-oneHotFeats-1-and-2-layer}) and achieve (near) perfect accuracy in the simpler cases. This is in line with our theoretical analysis (see Insight \ref{insight:GAT}), as the norms of a single-layer model are not constrained and thus the attention parameters have more freedom to change. 

Overall, the accuracy of GAT worsens drastically along two dimensions simultaneously: 
i) an increase in the depth of the model (due to increased unnecessary aggregation), and 
ii) an increase in the complexity of the task (due to an increase in the number of classes in an ER graph and consequently in node neighborhoods). In the interest of space, we defer results for GATE$_S$ to Fig. \ref{alphaDist-cora-oneHotFeats-gates} in Appendix \ref{appendix:additional-results} as aggregation patterns similar to GATE are observed.

Having established that GATE excels GAT in avoiding task-irrelevant neighborhood aggregation, it is also important to verify whether GATE can perform task-relevant neighborhood aggregation \textit{when} required, and \textit{as much as} required. We answer this question next by studying the behavior of GATE, in comparison to GAT, on a synthetically constructed neighbor-dependent learning problem.

\begin{table}[t]
\caption{ Neighbor-dependent learning: $k$ and $L$ denote the number of aggregation steps of the random GAT used for label generation and the number of layers of the evaluated network, respectively. Mean test accuracy $\pm 95\%$ confidence interval over 5 runs is reported. In all cases, a GATE variant outperforms the GAT variants. We further analyze one experimental run in detail in Table \ref{table:neighbor-dependent-learning-one-run} in Appendix \ref{appendix:additional-results}.  }
\label{table:nbr-dep-syn-avg5runs}
\centering
\begin{tblr}{
  width = \linewidth,
  colspec = {Q[4]Q[4]Q[152]Q[152]Q[152]Q[152]},
  cells = {c},
  cell{2}{1} = {r=3}{},
  cell{5}{1} = {r=3}{},
  cell{8}{1} = {r=3}{},
  hline{1,11} = {-}{0.08em},
  hline{2,5,8} = {-}{0.05em},
}
$k$ & $L$ & GAT$_S$ & GAT & GATE$_S$ & GATE\\
$1$ & $1$ & $93.6 \pm 1.3$ & $92.3 \pm 1.3$ & $\mathbf{96.4 \pm 0.7}$ & $93.5 \pm 1.3$\\
 & $2$ & $93.5 \pm 0.7$ & $92.7 \pm 2.7$ & $\mathbf{97.9 \pm 0.8}$ & $94.6 \pm 2.1$\\
 & $3$ & $88.2 \pm 4.9$ & $91.8 \pm 3.4$ & $92.1 \pm 4.6$ & $\mathbf{94.0 \pm 1.5}$\\
$2$ & $2$ & $90.4 \pm 1.3$ & $87.7 \pm 1.6$ & $\mathbf{93.8 \pm 0.5}$ & $88.7 \pm 2.5$\\
 & $3$ & $82.2 \pm 4.5$ & $88.9 \pm 2.1$ & $85.8 \pm 2.5$ & $\mathbf{93.4 \pm 3.3}$\\
 & $4$ & $84.0 \pm 5.0$ & $83.0 \pm 4.8$ & $\mathbf{89.2 \pm 2.3}$ & $87.8 \pm 2.4$\\
$3$ & $3$ & $84.3 \pm 3.2$ & $83.8 \pm 2.7$ & $87.5 \pm 1.8$ & $\mathbf{88.6 \pm 2.0}$\\
 & $4$ & $71.4 \pm 3.9$ & $75.9 \pm 7.6$ & $\mathbf{89.2 \pm 1.0}$ & $89.0 \pm 0.5$\\
 & $5$ & $80.2 \pm 4.8$ & $83.9 \pm 2.2$ & $86.1 \pm 0.8$ & $\mathbf{87.8 \pm 1.6}$
\end{tblr}

%%% double decimal place
% $k$ & $L$ & GAT$_S$ & GAT & GATE$_S$ & GATE\\
% $1$ & $1$ & $93.60 \pm 1.26$ & $92.32 \pm 1.27$ & $\mathbf{96.40 \pm 0.70}$ & $93.52 \pm 1.27$\\
%  & $2$ & $93.52 \pm 0.73$ & $92.72 \pm 2.69$ & $\mathbf{97.92 \pm 0.79}$ & $94.64 \pm 2.10$\\
%  & $3$ & $88.16 \pm 4.86$ & $91.76 \pm 3.35$ & $92.08 \pm 4.59$ & $\mathbf{94.00 \pm 1.54}$\\
% $2$ & $2$ & $90.40 \pm 1.30$ & $87.68 \pm 1.58$ & $\mathbf{93.84 \pm 0.51}$ & $88.72 \pm 2.50$\\
%  & $3$ & $82.16 \pm 4.45$ & $88.88 \pm 2.12$ & $85.76 \pm 2.54$ & $\mathbf{93.44 \pm 3.27}$\\
%  & $4$ & $84.00 \pm 5.00$ & $83.04 \pm 4.75$ & $\mathbf{89.20 \pm 2.33}$ & $87.76 \pm 2.36$\\
% $3$ & $3$ & $84.32 \pm 3.18$ & $83.84 \pm 2.69$ & $87.52 \pm 1.82$ & $\mathbf{88.64 \pm 2.00}$\\
%  & $4$ & $71.36 \pm 3.88$ & $75.92 \pm 7.63$ & $\mathbf{89.20 \pm 1.04}$ & $88.96 \pm 0.51$\\
%  & $5$ & $80.16 \pm 4.75$ & $83.92 \pm 2.16$ & $86.08 \pm 0.79$ & $\mathbf{87.84 \pm 1.59}$
\end{table}

% \begin{figure} [t]
% \centering
% \includegraphics[width=\linewidth]{Figures/alphaDistribution/alpha_ii_distribution_legend.png}\\
% \begin{subfigure}{.5\textwidth}
% \begin{subfigure}{.33\textwidth}
%   \centering
% \frame{\includegraphics[width=.99\linewidth]{Figures/alphaDistribution/randSynDataSparse_1aggr_1Layer_org.png}}
% \end{subfigure}%
% \begin{subfigure}{.66\textwidth}
%   \centering  
% \frame{\includegraphics[width=.98\linewidth]{Figures/alphaDistribution/randSynDataSparse_1aggr_2Layer_org.png}}
% \end{subfigure}
% \end{subfigure}\\
% \begin{subfigure}{.5\textwidth}
% \begin{subfigure}{.33\textwidth}
%   \centering
% \frame{\includegraphics[width=.99\linewidth]{Figures/alphaDistribution/randSynDataSparse_1aggr_1Layer_new.png}}
% \end{subfigure}%
% \begin{subfigure}{.66\textwidth}
%   \centering  
% \frame{\includegraphics[width=.98\linewidth]{Figures/alphaDistribution/randSynDataSparse_1aggr_2Layer_new.png}}
% \end{subfigure}
% \end{subfigure}
%   \caption{Distribution of $\alpha_{vv}$ against training epoch for the neighbor-dependent learning problem with $k=1$. Rows: GAT (top) and GATE(bottom). Columns (left to right): $1$, $2$, and $3$ layer models. While GAT is unable to switch off neighborhood aggregation, GATE allows most aggregation in mainly $1$ layer of the $1$ and $2$ layer models. Similarly, we observe for $k=3$, only $3$ layers of the $4$ and $5$ layer models perform neighborhood aggregation (see Fig. \ref{alpha-dist-randSynData-k3} in Appendix \ref{appendix:additional-results}).}
%   \label{alpha-dist-randSynData-k1-layers-one-and-two}
% \end{figure}

\paragraph{Learning neighbor-dependent node labels} To model this task, we generate an ER graph structure with $N=1000$ nodes and edge probability $p=0.01$. Input node features $\mathbf{h}_v^0 \in \mathbb{R}^d$ are sampled from a multivariate normal distribution $\mathcal{N}(\mathbf{0}_d,\mathbf{I}_d)$. For simplicity, $d=2$. 

This input graph $G$ is fed to a random GAT network $M_k$ with $k$ layers of width $d$. 
Note that this input graph $G$ has no self-loops on nodes (i.e. $v \notin \mathbb{N}(v)$).%, as is generally, though not necessarily, the case of graphs input to a GAT. 
The parameters of $M_k$ are initialized with the standard Xavier \citep{GlorotInit} initialization. 
Thus, for each node $v$, the node embedding output by $M_k$, $\mathbf{h}_v^{M_k}$ is effectively a function $f$ of the $k$-hop neighboring nodes of node $v$ represented by a random GAT network. 
Let $\mathbb{N}_k(v)$ denote the set of $k$-hop neighbors of $v$ and $v \notin \mathbb{N}_k(v)$. 

Finally, we run $K$-means clustering on the neighborhood aggregated representation of nodes $\mathbf{h}_v^{M_k}$ to divide nodes into $C$ clusters. For simplicity, we set $C=2$. 
This clustering serves as the node labels (i.e. $y_v = \arg_{c \in [C]}(v \in c)$ for our node classification task. 
Thus, the label $y_v$ of a node $v$ to be learned is highly dependent on the input features of the neighboring nodes $\mathbf{h}_u^0 \in \mathbb{N}_k(v)$ rather than the node's own input features $\mathbf{h}_v^0$. 

The generated input data and the real decision boundary for varying $k$ are shown in Fig. \ref{randSynData}.
Corresponding results in Table \ref{table:nbr-dep-syn-avg5runs} and Fig. \ref{alpha-dist-randSynData-k3} exhibit that GATE can better detect the amount of necessary neighborhood aggregation than GAT. 
However, this task is more challenging than the previous one, and GATE too can not achieve perfect $100\%$ test accuracy.
This could be attributed to data points close to the real decision boundary which is not crisply defined (see Fig. \ref{randSynData}).

\subsection{Real-World Data}

To demonstrate that the ability of GATE to switch-off neighborhood aggregation has real application relevance, we evaluate GATE on relatively large-scale real-world node classification tasks, namely on five heterophilic benchmark datasets \citep{criticalHet}  (see Table \ref{real-het-results}) and three OGB datasets \citep{ogb} (see Table \ref{real-ogb-results}). We defer results and discussion on five small-scale datasets with varying homophily levels to Table \ref{table:real-small-data} in Appendix \ref{appendix:additional-results}. 
To analyze the potential benefits of combining MLP and GAT layers in GATE, we compare its behavior with GAT and MLP. 
We argue that the better performance of GATE, by a large margin in most cases, can be attributed to down-weighting unrelated neighbors, leveraging deeper non-linear feature transformations, and reducing over-smoothing. 

While we focus our exposition on the neighborhood aggregation perspective of GATs, 
% we do not compare extensively with SOTA methods designed specifically for heterophilic datasets. However, we do
we also consider the FAGCN architecture \cite{fagcn}, which relies on a similar attention mechanism and, in theory, could switch off neighborhood aggregation when positive and negative contributions of neighbors cancel out.
In contrast to GATE, it requires tuning a hyperparameter $\epsilon$, which controls the contribution of raw node features to each layer.
%FAGCN explicitly feeds raw node features to every layer in the network. 
%As self-loops are removed in the input graph, the contribution of the raw node features in each layer is determined by a hyper-parameter $\epsilon$ that likely is sensitive to the task and requires tuning which GATE avoids. 
Furthermore, on our synthetic tasks, we find that, like GAT, FAGCN is also unable to limit neighborhood contribution. We also provide a detailed qualitative and quantitative discussion comparing GATE and FAGCN using the synthetic testbed in Appendix \ref{appendix:additional-results}.

\begin{figure*} [t]
\centering
\includegraphics[width=.75\linewidth]{Figures/alphaDistribution/alpha_ii_distribution_legend.png}\\
\begin{subfigure}{.5\textwidth}
  \centering
\frame{\includegraphics[width=.99\linewidth]{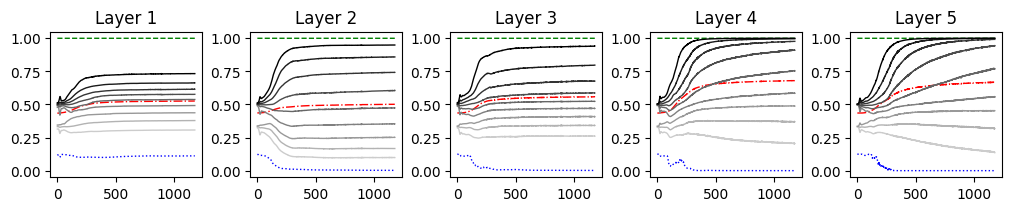}}
\caption{roman-empire, GAT: $28.96\%$ test accuracy.}
\end{subfigure}%
\begin{subfigure}{.5\textwidth}
  \centering  
\frame{\includegraphics[width=.99\linewidth]{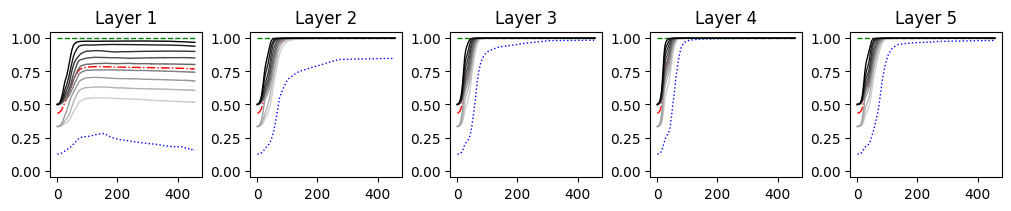}}
\caption{roman-empire, GATE: $75.94\%$ test accuracy.}
\end{subfigure}\\
\begin{subfigure}{.5\textwidth}
  \centering
\frame{\includegraphics[width=.99\linewidth]{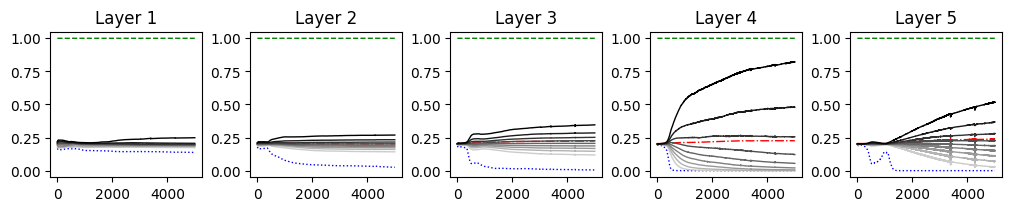}}
\caption{Minesweeper, GAT: $50.50\%$ test AUROC.}
\end{subfigure}%
\begin{subfigure}{.5\textwidth}
  \centering  
\frame{\includegraphics[width=.99\linewidth]{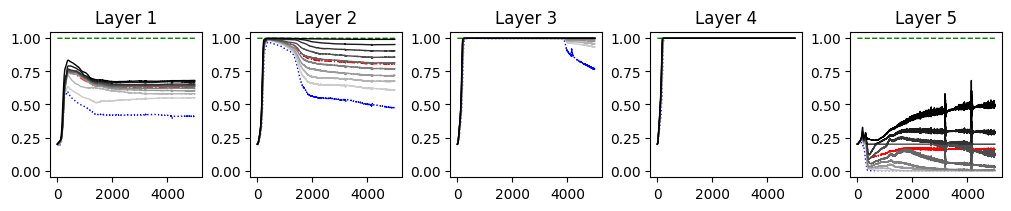}}
\caption{Minesweeper, GATE: $67.57\%$ test AUROC.}
\end{subfigure}\\
% \begin{subfigure}{.5\textwidth}
%   \centering
% \frame{\includegraphics[width=.99\linewidth]{Figures/alphaDistribution/tolokers_10layer_org.png}}
% \caption{Tolokers, GAT:}
% \end{subfigure}%
% \begin{subfigure}{.5\textwidth}
%   \centering  
% \frame{\includegraphics[width=.99\linewidth]{Figures/alphaDistribution/tolokers_10layer_new.png}}
% \caption{Tolokers, GATE:}
% \end{subfigure}\\
  \caption{Distribution of $\alpha_{vv}$ against training epoch for a run of 5 layer networks on real-world heterophilic tasks. As expected, GAT is unable to significantly vary neighborhood aggregation whereas GATE exhibits varying aggregation patterns across layers and tasks. These could be interpreted to indicate the inherent importance of raw node features relative to their neighborhoods for a given task. We defer the plots of the 10-layer models for tolokers dataset to Fig. \ref{alpha-dist-gate-tolokers} in Appendix \ref{appendix:additional-results}. }
  \label{alpha-dist-gate-het}
\end{figure*}

Next, we analyze GATE's ability to mix MLP and GAT layers.
To this end, we evaluate a GNN architecture constructed by alternately placing GAT and MLP layers in the network that we denote by MLP$_{+GAT}$ on various heterophilic tasks. The purpose of this experiment is twofold. Firstly, we observe in Table \ref{real-het-results} that MLP$_{+GAT}$ outperforms both GAT and MLP in most cases. This highlights the benefit of only performing non-linear transformations on raw or aggregated neighborhood features without immediate further neighborhood aggregation to learn potentially more complex features. Secondly, we find GATE to outperform MLP$_{+GAT}$ (see Table \ref{real-het-results}). This illustrates that rigidly embedding MLP layers in a GNN with arbitrary predefined roles is not ideal as the appropriate degree and placement of neighborhood aggregation is unknown a-priori. In contrast, GATE offers more flexibility to learn intricate combinations of GNN layers and nonlinear feature transformations that define more adequate models for a given task, as exemplified in Fig. \ref{alpha-dist-gate-het}. 

%  We verify this by observing the neighborhood aggregation patterns of GATE in Fig. \ref{alpha-dist-gate-real-heterophilic} that shows neighborhood aggregation is switched off in some layers by GATE.  
% As expected, no layers switch off neighborhood aggregation in the GAT network (see Fig. (\ref{alpha-dist-gat-real-heterophilic}) in Appendix \ref{appendix:additional-results}). 

 \begin{table*}[h]
\caption{ We report mean test accuracy $\pm95\%$ confidence interval for roman-empire and amazon-ratings and AUC-ROC for the other three datasets over the standard $10$ splits, following  \citep{criticalHet}. All architectures were run with networks of depth 5 and 10 layers. The better performance for each architecture is shown with the number of network layers used in parentheses. GATE outperforms GAT and other baselines on all datasets, mostly by a significant margin.}
\label{real-het-results}
\centering
\begin{tblr}{
  width = \linewidth,
  colspec = {Q[80]Q[154]Q[154]Q[154]Q[154]Q[154]},
  cell{1}{2} = {c},
  cell{1}{3} = {c},
  cell{1}{4} = {c},
  cell{1}{5} = {c},
  cell{1}{6} = {c},
  cell{2}{1} = {c},
  cell{3}{1} = {c},
  cell{4}{1} = {c},
  cell{5}{1} = {c},
  cell{6}{1} = {c},
  hline{1,7} = {-}{0.08em},
  hline{2} = {-}{0.05em},
}
        & roman-empire           & amazon-ratings          & questions               & minesweeper             & tolokers                \\
GAT     & $26.10 \pm 1.25 \:(5)$ & $45.58 \pm 0.41 \:(10)$ & $57.72 \pm 1.58 \:(5)$  & $50.83 \pm 0.41 \:(5)$  & $63.57 \pm 1.03 \:(10)$ \\
MLP     & $65.12 \pm 0.25 \:(5)$ & $43.26 \pm 0.34 \:(5)$  & $59.44 \pm 0.94 \:(10)$ & $50.74 \pm 0.56 \:(5)$  & $62.67 \pm 1.06 \:(10)$ \\
MLP$_{+GAT}$ & $70.83 \pm 0.39 \:(5)$ & $45.25 \pm 0.17 \:(10)$ & $59.12 \pm 1.57 \:(10)$ & $60.07 \pm 1.11 \:(5)$  & $65.85 \pm 0.64 \:(10)$ \\
FAGCN   & $67.55 \pm 0.81 \:(5)$ & $42.85 \pm 0.83 \:(10)$ & $60.38 \pm 1.21 \:(5)$  & $63.38 \pm 0.91 \:(10)$ & $60.89 \pm 1.12 \:(5)$  \\
GATE    & $\mathbf{75.55 \pm 0.30} \:(5)$ & $\mathbf{45.73 \pm 0.24} \:(10)$ & $\mathbf{62.95 \pm 0.71} \:(5)$  & $\mathbf{66.14 \pm 1.57} \:(5)$  & $\mathbf{66.63 \pm 1.15} \:(10)$ 
\end{tblr}

\end{table*}

%\paragraph{Interpretable neighborhood aggregation}
The distributions of $\alpha_{vv}$ in Fig. \ref{alpha-dist-gate-het} across layers in GATE reveal information about the relative importance of node feature and graph structure at the node level, which allows us to analyze the question to which degree graph information is helpful for a task. For example, in Fig. \ref{alpha-dist-gate-het}, we observe that the $\alpha_{vv}$ values are mostly lower in the minesweeper dataset than the roman-empire dataset. This indicates that aggregation, particularly over the input node features and the final layer's learned representations, is more beneficial compared to the node's own features for the minesweeper dataset. On the other hand, for roman-empire, the model has a higher preference to utilize features of the node itself (as most values of $\alpha_{vv}$ approach 1) over features of the neighbors. This aligns with the homophily levels, 0.05 and 0.68, of the roman-empire and minesweeper datasets, respectively. A similar analysis for datasets Texas and Actor can be found in Fig. \ref{alpha-dist-gate-real-heterophilic} in Appendix \ref{appendix:additional-results}.
%can be interpreted as importance of input features of nodes relative to their neighborhood. 

We also observe in Fig. \ref{alpha-dist-gate-het} that when neighborhood aggregation takes place, the level of aggregation across all nodes, as indicated by the shape of $\alpha_{vv}$ distribution, varies over network layers. This is expected as different nodes need different levels of aggregation depending on where they are situated in the graph topology. For example, peripheral nodes would require more aggregation than central nodes to obtain a similar amount of information. %Obviously, the distribution shape of $\alpha_{vv}$ also varies across datasets with different topological structures.
Therefore, as already observed with purposefully constructed synthetic data, GATE offers a more interpretable model than GAT in a real-world setting. 

While we focus our evaluation in Table \ref{real-het-results} on comparison with the most relevant baselines such as attention-based architectures, we next present a more extensive comparison with $14$ other baseline architectures in Table \ref{real-heterophilic-baselines}.
For the results reported in Table \ref{real-het-results}, we conduct experiments in a simple setting without additional elements that may impact the performance such as skip connections, normalization, etc., to isolate the effect of the architecture and evaluate solely the impact of GATE's ability to switch off neighborhood aggregation on real-world data. 
However, for the results in Table \ref{real-heterophilic-baselines}, we adopt the original codebase of \cite{criticalHet}, which utilizes such elements to evaluate the performance of baseline GNNs and architectures specifically designed for heterophilic datasets.
%To evaluate solely the impact of GATE's ability to switch off neighborhood aggregation on real-world data, we have focused in Table \ref{table:real-small-data} on a comparison with most relevant baselines such as attention-based architectures, particularly those that claim to have the ability to switch off neighborhood aggregation such as FAGCN, in a simple setting without additional elements that may impact the performance such as skip connections, normalization, etc., to isolate the effect of the architecture.  
We evaluate GATE in the same settings optimized for their experiments. For easy comparison, we replicate their results from Table 4 in \cite{criticalHet}.

We observe in Table \ref{real-heterophilic-baselines} that while GATE outperforms GAT (and other baselines) significantly, GATE has comparable performance to GAT-sep, a variant of GAT, despite GATE being more parameter efficient by an order of magnitude. More specifically, GAT-sep and GATE introduce $d^2$ and $d$ additional parameters, respectively, in a layer. By correspondingly adapting GATE, we find GATE-sep to achieve the best performance in most cases. Therefore, additional techniques generally employed to boost performance are compatible and complementary to GATE. 

 \begin{table*}[t]
\caption{An extensive comparison of GATE with baseline GNNs using the experimental setup of \cite{criticalHet}. Accuracy is reported for roman-empire and amazon-ratings, and ROC AUC is reported for the remaining three datasets.}
\label{real-heterophilic-baselines}
\centering
\begin{tblr}{
  width = \linewidth,
  colspec = {Q[110]Q[150]Q[150]Q[150]Q[150]Q[150]},
  row{1} = {c},
  cell{2}{2} = {c},
  cell{2}{3} = {c},
  cell{2}{4} = {c},
  cell{2}{5} = {c},
  cell{2}{6} = {c},
  cell{3}{2} = {c},
  cell{3}{3} = {c},
  cell{3}{4} = {c},
  cell{3}{5} = {c},
  cell{3}{6} = {c},
  cell{4}{2} = {c},
  cell{4}{3} = {c},
  cell{4}{4} = {c},
  cell{4}{5} = {c},
  cell{4}{6} = {c},
  cell{5}{2} = {c},
  cell{5}{3} = {c},
  cell{5}{4} = {c},
  cell{5}{5} = {c},
  cell{5}{6} = {c},
  cell{6}{2} = {c},
  cell{6}{3} = {c},
  cell{6}{4} = {c},
  cell{6}{5} = {c},
  cell{6}{6} = {c},
  cell{7}{2} = {c},
  cell{7}{3} = {c},
  cell{7}{4} = {c},
  cell{7}{5} = {c},
  cell{7}{6} = {c},
  cell{8}{2} = {c},
  cell{8}{3} = {c},
  cell{8}{4} = {c},
  cell{8}{5} = {c},
  cell{8}{6} = {c},
  cell{9}{2} = {c},
  cell{9}{3} = {c},
  cell{9}{4} = {c},
  cell{9}{5} = {c},
  cell{9}{6} = {c},
  cell{10}{2} = {c},
  cell{10}{3} = {c},
  cell{10}{4} = {c},
  cell{10}{5} = {c},
  cell{10}{6} = {c},
  cell{11}{2} = {c},
  cell{11}{3} = {c},
  cell{11}{4} = {c},
  cell{11}{5} = {c},
  cell{11}{6} = {c},
  cell{12}{2} = {c},
  cell{12}{3} = {c},
  cell{12}{4} = {c},
  cell{12}{5} = {c},
  cell{12}{6} = {c},
  cell{13}{2} = {c},
  cell{13}{3} = {c},
  cell{13}{4} = {c},
  cell{13}{5} = {c},
  cell{13}{6} = {c},
  cell{14}{2} = {c},
  cell{14}{3} = {c},
  cell{14}{4} = {c},
  cell{14}{5} = {c},
  cell{14}{6} = {c},
  cell{15}{2} = {c},
  cell{15}{3} = {c},
  cell{15}{4} = {c},
  cell{15}{5} = {c},
  cell{15}{6} = {c},
  cell{16}{2} = {c},
  cell{16}{3} = {c},
  cell{16}{4} = {c},
  cell{16}{5} = {c},
  cell{16}{6} = {c},
  cell{17}{2} = {c},
  cell{17}{3} = {c},
  cell{17}{4} = {c},
  cell{17}{5} = {c},
  cell{17}{6} = {c},
  cell{18}{2} = {c},
  cell{18}{3} = {c},
  cell{18}{4} = {c},
  cell{18}{5} = {c},
  cell{18}{6} = {c},
  cell{19}{2} = {c},
  cell{19}{3} = {c},
  cell{19}{4} = {c},
  cell{19}{5} = {c},
  cell{19}{6} = {c},
  cell{20}{2} = {c},
  cell{20}{3} = {c},
  cell{20}{4} = {c},
  cell{20}{5} = {c},
  cell{20}{6} = {c},
  hline{1-2,4,6,10,18,21} = {-}{},
}
           & roman-empire              & amazon-ratings            & minesweeper               & tolokers                  & questions                 \\
GATE       & 89.51 $\pm$ 0.49          & 52.49 $\pm$ 0.46          & 92.82 $\pm$ 0.90          & 84.62 $\pm$ 0.69          & 78.46 $\pm$ 1.17          \\
GAT        & 80.87 $\pm$ 0.30          & 49.09 $\pm$ 0.63          & 92.01 $\pm$ 0.68          & 83.70 $\pm$ 0.47          & 77.43 $\pm$ 1.20          \\
GATE-sep   & \textbf{89.78 $\pm$ 0.54} & \textbf{54.51 $\pm$ 0.38} & \textbf{94.18 $\pm$ 0.43} & \textbf{84.48 $\pm$ 0.57} & 78.20 $\pm$ 1.00          \\
GAT-sep    & 88.75 $\pm$ 0.41          & 52.70 $\pm$ 0.62          & 93.91 $\pm$ 0.35          & 83.78 $\pm$ 0.43          & 76.79 $\pm$ 0.71          \\
GT         & 86.51 $\pm$ 0.73          & 51.17 $\pm$ 0.66          & 91.85 $\pm$ 0.76          & 83.23 $\pm$ 0.64          & 77.95 $\pm$ 0.68          \\
GT-sep     & 87.32 $\pm$ 0.39          & 52.18 $\pm$ 0.80          & 92.29 $\pm$ 0.47          & 82.52 $\pm$ 0.92          & 78.05 $\pm$ 0.93          \\
GCN        & 73.69 $\pm$ 0.74          & 48.70 $\pm$ 0.63          & 89.75 $\pm$ 0.52          & 83.64 $\pm$ 0.67          & 76.09 $\pm$ 1.27          \\
SAGE       & 85.74 $\pm$ 0.67          & 53.63 $\pm$ 0.39          & 93.51 $\pm$ 0.57          & 82.43 $\pm$ 0.44          & 76.44 $\pm$ 0.62          \\
H$_2$GCN      & 60.11 $\pm$ 0.52          & 36.47 $\pm$ 0.23          & 89.71 $\pm$ 0.31          & 73.35 $\pm$ 1.01          & 63.59 $\pm$ 1.46          \\
CPGNN      & 63.96 $\pm$ 0.62          & 39.79 $\pm$ 0.77          & 52.03 $\pm$ 5.46          & 73.36 $\pm$ 1.01          & 65.96 $\pm$ 1.95          \\
GPR-GNN    & 64.85 $\pm$ 0.27          & 44.88 $\pm$ 0.34          & 86.24 $\pm$ 0.61          & 72.94 $\pm$ 0.97          & 55.48 $\pm$ 0.91          \\
FSGNN      & 79.92 $\pm$ 0.56          & 52.74 $\pm$ 0.83          & 90.08 $\pm$ 0.70          & 82.76 $\pm$ 0.61          & \textbf{78.86 $\pm$ 0.92} \\
GloGNN     & 59.63 $\pm$ 0.69          & 36.89 $\pm$ 0.14          & 51.08 $\pm$ 1.23          & 73.39 $\pm$ 1.17          & 65.74 $\pm$ 1.19          \\
FAGCN      & 65.22 $\pm$ 0.56          & 44.12 $\pm$ 0.30          & 88.17 $\pm$ 0.73          & 77.75 $\pm$ 1.05          & 77.24 $\pm$ 1.26          \\
GBK-GNN    & 74.57 $\pm$ 0.47          & 45.98 $\pm$ 0.71          & 90.85 $\pm$ 0.58          & 81.01 $\pm$ 0.67          & 74.47 $\pm$ 0.86          \\
JacobiConv & 71.14 $\pm$ 0.42          & 43.55 $\pm$ 0.48          & 89.66 $\pm$ 0.40          & 68.66 $\pm$ 0.65          & 73.88 $\pm$ 1.16          \\
ResNet     & 65.88 $\pm$ 0.38          & 45.90 $\pm$ 0.52          & 50.89 $\pm$ 1.39          & 72.95 $\pm$ 1.06          & 70.34 $\pm$ 0.76          \\
ResNet+SGC & 73.90 $\pm$ 0.51          & 50.66 $\pm$ 0.48          & 70.88 $\pm$ 0.90          & 80.70 $\pm$ 0.97          & 75.81 $\pm$ 0.96          \\
ResNet+adj & 52.25 $\pm$ 0.40          & 51.83 $\pm$ 0.57          & 50.42 $\pm$ 0.83          & 78.78 $\pm$ 1.11          & 75.77 $\pm$ 1.24          
\end{tblr}
\end{table*}

%\input{Tables/table3}

% \begin{figure*} [t]
% \centering
% \includegraphics[width=.75\linewidth]{Figures/alphaDistribution/alpha_ii_distribution_legend.png}\\
% \begin{subfigure}{.275\textwidth}
%   \centering
% \frame{\includegraphics[width=.99\linewidth]{Figures/alphaDistribution/texas_2layer_new.png}}
% %\caption{Texas, GATE: 67.6@7}
% \end{subfigure}%
% \begin{subfigure}{.675\textwidth}
%   \centering  
% \frame{\includegraphics[width=.99\linewidth]{Figures/alphaDistribution/texas_5layer_new.png}}
% %\caption{Texas, GATE: 67.6@18}
% \end{subfigure}\\
% \begin{subfigure}{.275\textwidth}
%   \centering
% \frame{\includegraphics[width=.99\linewidth]{Figures/alphaDistribution/actor_2layer_new.png}}
% %\caption{Actor, GATE: 31.6@34}
% \end{subfigure}%
% \begin{subfigure}{.675\textwidth}
%   \centering  
% \frame{\includegraphics[width=.99\linewidth]{Figures/alphaDistribution/actor_5layer_new.png}}
% %\caption{Actor, GATE: 29.2@948 }
% \end{subfigure}\\
%   \caption{Distribution of $\alpha_{vv}$, against training epoch of 2-layer (left) and 5-layer (right) GATE networks for heterophilic datasets Texas (top) and Actor (bottom), across layers could be interpreted to indicate the inherent importance of raw node features relative to their neighborhoods.}
%   \label{alpha-dist-gate-real-heterophilic}
% \end{figure*}

 \begin{table}[t]
\caption{Mean test accuracy $\pm95\%$ confidence interval (and number of network layers). We replicate the results for GAT reported by \citep{gat}. GATE leverages deeper networks to substantially outperform GAT.}
\label{real-ogb-results}
\centering
\begin{tblr}{
  width = \linewidth,
  colspec = {Q[200]Q[370]Q[370]},
  row{1} = {c},
  cell{2}{1} = {c},
  cell{3}{1} = {c},
  cell{4}{1} = {c},
  hline{1,5} = {-}{0.08em},
  hline{2} = {-}{0.05em},
}
 OGB-            & GAT                    & GATE                    \\
arxiv    & $71.87 \pm 0.16$ $(3)$ & $\mathbf{79.57 \pm 0.84}$ $(12)$ \\
products & $80.63 \pm 0.46$ $(3)$ & $\mathbf{86.24 \pm 1.01}$ $(8)$  \\
mag      & $32.61 \pm 0.29$ $(2)$ & $\mathbf{35.29 \pm .36}$ $(5) $  
\end{tblr}
\end{table}

GATE's ability to benefit from depth in terms of generalization is demonstrated on OGB datasets (see Table \ref{real-ogb-results}). In particular, GATE improves the SOTA test accuracy $(78.03\%)$  on the arxiv dataset achieved by a model using embeddings learned by a language model instead of raw node features\citep{simteg}, as reported on the OGB leaderboard. 
While the better performance of deeper models with limited neighborhood aggregation in certain layers indicates reduced over-smoothing, we also verify this insight quantitatively (see Table \ref {smoothness_DE} in Appendix \ref{appendix:additional-results}). 

Our experimental code is available at \url{https://github.com/RelationalML/GATE.git}.

 %We observe that the advantage of GATE is greater in heterophilic tasks that are more susceptible to unnecessary neighborhood aggregations than homophilic tasks, where, while the performance gain is limited, the greater effect is a reduction in over-smoothing with network depth. 

\section{Conclusion}
We experimentally illustrate a structural limitation %failure point
of GAT that disables the architecture, in practice, to switch off task-irrelevant neighborhood aggregation. 
This obstructs GAT from achieving its intended potential.
Based on insights from an existing conservation law of gradient flow dynamics in GAT, we have explained the source of this problem.
To verify that we have identified the correct issue, we resolve it with a modification of GAT, which we call GATE, and derive the corresponding modified conservation law. 
GATE holds multiple advantages over GAT, as it can leverage the benefits of depth as in MLPs, offer interpretable, learned self-attention coefficients, and adapt the model to the necessary degree of neighborhood aggregation for a given task. 
We verify this on multiple synthetic and real-world tasks, where GATE significantly outperforms GAT and also achieves a new SOTA test accuracy on the OGB-arxiv dataset.
Therefore, we argue that GATE is a suitable candidate to answer highly debated questions related to the importance of a given graph structure for standard tasks.

 \section*{Acknowledgements}
We gratefully acknowledge funding from the European Research Council (ERC) under the Horizon Europe Framework Programme (HORIZON) for proposal number 101116395 SPARSE-ML.
% \textbf{Do not} include acknowledgements in the initial version of
% the paper submitted for blind review.

% If a paper is accepted, the final camera-ready version can (and
% probably should) include acknowledgements. In this case, please
% place such acknowledgements in an unnumbered section at the
% end of the paper. Typically, this will include thanks to reviewers
% who gave useful comments, to colleagues who contributed to the ideas,
% and to funding agencies and corporate sponsors that provided financial


\section*{Impact Statement}
This paper presents work whose goal is to advance the field of Machine Learning. There are many potential societal consequences of our work, none of which we feel must be specifically highlighted here.

\bibliography{references}

\begin{thebibliography}{53}
\providecommand{\natexlab}[1]{#1}
\providecommand{\url}[1]{\texttt{#1}}
\expandafter\ifx\csname urlstyle\endcsname\relax
  \providecommand{\doi}[1]{doi: #1}\else
  \providecommand{\doi}{doi: \begingroup \urlstyle{rm}\Url}\fi

\bibitem[Alon \& Yahav(2021)Alon and Yahav]{oversquashing}
Alon, U. and Yahav, E.
\newblock On the bottleneck of graph neural networks and its practical implications.
\newblock In \emph{International Conference on Learning Representations}, 2021.

\bibitem[Bian et~al.(2020)Bian, Xiao, Xu, Zhao, Huang, Rong, and Huang]{Bian2020Rumor}
Bian, T., Xiao, X., Xu, T., Zhao, P., Huang, W., Rong, Y., and Huang, J.
\newblock Rumor detection on social media with bi-directional graph convolutional networks.
\newblock In \emph{AAAI Conference on Artificial Intelligence}, 2020.

\bibitem[Bo et~al.(2021)Bo, Wang, Shi, and Shen]{fagcn}
Bo, D., Wang, X., Shi, C., and Shen, H.
\newblock Beyond low-frequency information in graph convolutional networks.
\newblock In \emph{AAAI Conference on Artificial Intelligence}, 2021.

\bibitem[Brody et~al.(2022)Brody, Alon, and Yahav]{gat}
Brody, S., Alon, U., and Yahav, E.
\newblock How attentive are graph attention networks?
\newblock In \emph{International Conference on Learning Representations}, 2022.

\bibitem[Burkholz \& Dubatovka(2019)Burkholz and Dubatovka]{dyniso}
Burkholz, R. and Dubatovka, A.
\newblock Initialization of {ReLUs} for dynamical isometry.
\newblock In \emph{Advances in Neural Information Processing Systems}, volume~32, 2019.

\bibitem[Cai \& Wang(2020)Cai and Wang]{cai2020note}
Cai, C. and Wang, Y.
\newblock A note on over-smoothing for graph neural networks.
\newblock In \emph{Graph Representation Learning Workshop, International Conference on Machine Learning}, 2020.

\bibitem[Cai et~al.(2021)Cai, Luo, Xu, He, Liu, and Wang]{graphnorm}
Cai, T., Luo, S., Xu, K., He, D., Liu, T.-Y., and Wang, L.
\newblock Graphnorm: A principled approach to accelerating graph neural network training.
\newblock In \emph{International Conference on Machine Learning}, 2021.

\bibitem[Chen et~al.(2020)Chen, Wei, Huang, Ding, and Li]{chen2020simple}
Chen, M., Wei, Z., Huang, Z., Ding, B., and Li, Y.
\newblock Simple and deep graph convolutional networks.
\newblock In \emph{International Conference on Machine Learning}, 2020.

\bibitem[Cong et~al.(2021)Cong, Ramezani, and Mahdavi]{depthBenefitsGNNs}
Cong, W., Ramezani, M., and Mahdavi, M.
\newblock On provable benefits of depth in training graph convolutional networks.
\newblock In \emph{Advances in Neural Information Processing Systems}, 2021.

\bibitem[Deac et~al.(2022)Deac, Lackenby, and Veličković]{deac2022expander}
Deac, A., Lackenby, M., and Veličković, P.
\newblock Expander graph propagation.
\newblock In \emph{Learning on Graphs Conference}, 2022.

\bibitem[Duan et~al.(2023)Duan, Liu, Chua, Yan, Ooi, Xie, and He]{simteg}
Duan, K., Liu, Q., Chua, T.-S., Yan, S., Ooi, W.~T., Xie, Q., and He, J.
\newblock Simteg: A frustratingly simple approach improves textual graph learning, 2023.

\bibitem[Eliasof et~al.(2023)Eliasof, Ruthotto, and Treister]{wGAT}
Eliasof, M., Ruthotto, L., and Treister, E.
\newblock Improving graph neural networks with learnable propagation operators.
\newblock In \emph{International Conference on Machine Learning}, 2023.

\bibitem[Errica et~al.(2023)Errica, Christiansen, Zaverkin, Maruyama, Niepert, and Alesiani]{AMP}
Errica, F., Christiansen, H., Zaverkin, V., Maruyama, T., Niepert, M., and Alesiani, F.
\newblock Adaptive message passing: A general framework to mitigate oversmoothing, oversquashing, and underreaching, 2023.

\bibitem[Finkelshtein et~al.(2023)Finkelshtein, Huang, Bronstein, and İsmail~İlkan Ceylan]{coopGNNs}
Finkelshtein, B., Huang, X., Bronstein, M., and İsmail~İlkan Ceylan.
\newblock Cooperative graph neural networks, 2023.

\bibitem[Fountoulakis et~al.(2023)Fountoulakis, Levi, Yang, Baranwal, and Jagannath]{fountoulakis2023graph}
Fountoulakis, K., Levi, A., Yang, S., Baranwal, A., and Jagannath, A.
\newblock Graph attention retrospective.
\newblock In \emph{Journal of Machine Learning Research}, 2023.

\bibitem[Franceschi et~al.(2020)Franceschi, Niepert, Pontil, and He]{franceschi2020learning}
Franceschi, L., Niepert, M., Pontil, M., and He, X.
\newblock Learning discrete structures for graph neural networks.
\newblock In \emph{International Conference on Machine Learning}, 2020.

\bibitem[Gasteiger et~al.(2019)Gasteiger, Bojchevski, and Günnemann]{appnp}
Gasteiger, J., Bojchevski, A., and Günnemann, S.
\newblock Predict then propagate: Graph neural networks meet personalized pagerank.
\newblock In \emph{International Conference on Learning Representations}, 2019.

\bibitem[Glorot \& Bengio(2010)Glorot and Bengio]{GlorotInit}
Glorot, X. and Bengio, Y.
\newblock Understanding the difficulty of training deep feedforward neural networks.
\newblock In \emph{International Conference on Artificial Intelligence and Statistics}, volume~9, pp.\  249--256, May 2010.

\bibitem[Gomes et~al.(2022)Gomes, Ruelens, Efthymiadis, Nowe, and Vrancx]{gomes2022when}
Gomes, D., Ruelens, F., Efthymiadis, K., Nowe, A., and Vrancx, P.
\newblock When are graph neural networks better than structure agnostic methods?
\newblock In \emph{Neural Information Processing Systems Workshop ICBINB}, 2022.

\bibitem[Gori et~al.(2005)Gori, Monfardini, and Scarselli]{Gori2005NewModel}
Gori, M., Monfardini, G., and Scarselli, F.
\newblock A new model for learnig in graph domains.
\newblock In \emph{IEEE International Joint Conference on Neural Networks}, 2005.

\bibitem[Hamaguchi et~al.(2017)Hamaguchi, Oiwa, Shimbo, and Matsumoto]{hamaguchi2017knowledge}
Hamaguchi, T., Oiwa, H., Shimbo, M., and Matsumoto, Y.
\newblock Knowledge transfer for out-of-knowledge-base entities : A graph neural network approach.
\newblock In \emph{International Joint Conference on Artificial Intelligence}, 2017.

\bibitem[Hamilton et~al.(2018)Hamilton, Ying, and Leskovec]{gcnHamilton}
Hamilton, W.~L., Ying, R., and Leskovec, J.
\newblock Inductive representation learning on large graphs.
\newblock In \emph{Advances in Neural Information Processing Systems}, 2018.

\bibitem[Hu et~al.(2021)Hu, Fey, Zitnik, Dong, Ren, Liu, Catasta, and Leskovec]{ogb}
Hu, W., Fey, M., Zitnik, M., Dong, Y., Ren, H., Liu, B., Catasta, M., and Leskovec, J.
\newblock Open graph benchmark: Datasets for machine learning on graphs, 2021.

\bibitem[Kearnes et~al.(2016)Kearnes, McCloskey, Berndl, Pande, and Riley]{Kearnes2016Molecular}
Kearnes, S., McCloskey, K., Berndl, M., Pande, V., and Riley, P.
\newblock Molecular graph convolutions: Moving beyond fingerprints.
\newblock In \emph{Journal of Computer-Aided Molecular Design}, 2016.

\bibitem[Kim \& Oh(2021)Kim and Oh]{superGAT}
Kim, D. and Oh, A.
\newblock How to find your friendly neighborhood: Graph attention design with self-supervision.
\newblock In \emph{International Conference on Learning Representations}, 2021.

\bibitem[Kipf \& Welling(2017)Kipf and Welling]{gcnKipf}
Kipf, T.~N. and Welling, M.
\newblock Semi-supervised classification with graph convolutional networks.
\newblock In \emph{International Conference on Learning Representations}, 2017.

\bibitem[Lee et~al.(2023)Lee, Bu, Yoo, and Shin]{lee2023deep}
Lee, S.~Y., Bu, F., Yoo, J., and Shin, K.
\newblock Towards deep attention in graph neural networks: Problems and remedies.
\newblock In \emph{International Conference on Machine Learning}, 2023.

\bibitem[Li et~al.(2018)Li, Han, and Wu]{oversmoothing}
Li, Q., Han, Z., and Wu, X.-M.
\newblock Deeper insights into graph convolutional networks for semi-supervised learning.
\newblock In \emph{AAAI Conference on Artificial Intelligence}, 2018.

\bibitem[Liu et~al.(2020)Liu, Gao, and Ji]{dagnn}
Liu, M., Gao, H., and Ji, S.
\newblock Towards deeper graph neural networks.
\newblock In \emph{SIGKDD International Conference on Knowledge Discovery and Data Mining}, 2020.

\bibitem[Ma et~al.(2022)Ma, Liu, Shah, and Tang]{ma2023homophily}
Ma, Y., Liu, X., Shah, N., and Tang, J.
\newblock Is homophily a necessity for graph neural networks?
\newblock In \emph{International Conference on Learning Representations}, 2022.

\bibitem[Ma et~al.(2023)Ma, Liu, Shah, and Tang]{Ma2022homophily}
Ma, Y., Liu, X., Shah, N., and Tang, J.
\newblock Is homophily a necessity for graph neural networks?
\newblock In \emph{International Conference on Learning Representations}, 2023.

\bibitem[Monken et~al.(2021)Monken, Haberkorn, Gopinatha, Freeman, and Batarseh]{monken2021intTrade}
Monken, A., Haberkorn, F., Gopinatha, M., Freeman, L., and Batarseh, F.~A.
\newblock Graph neural networks for modeling causality in international trade.
\newblock In \emph{AAAI Conference on Artificial Intelligence}, 2021.

\bibitem[Mustafa \& Burkholz(2023)Mustafa and Burkholz]{balanceGATs}
Mustafa, N. and Burkholz, R.
\newblock Are {GATS} out of balance?
\newblock In \emph{Advances in Neural Information Processing Systems}, 2023.

\bibitem[Papp et~al.(2021)Papp, Martinkus, Faber, and Wattenhofer]{dropgnn}
Papp, P.~A., Martinkus, K., Faber, L., and Wattenhofer, R.
\newblock Dropgnn: Random dropouts increase the expressiveness of graph neural networks.
\newblock In \emph{Advances in Neural Information Processing Systems}, 2021.

\bibitem[Pei et~al.(2020)Pei, Wei, Chang, Lei, and Yang]{geomgcn}
Pei, H., Wei, B., Chang, K. C.-C., Lei, Y., and Yang, B.
\newblock Geom-gcn: Geometric graph convolutional networks.
\newblock In \emph{International Conference on Learning Representations}, 2020.

\bibitem[Platonov et~al.(2023)Platonov, Kuznedelev, Diskin, Babenko, and Prokhorenkova]{criticalHet}
Platonov, O., Kuznedelev, D., Diskin, M., Babenko, A., and Prokhorenkova, L.
\newblock A critical look at the evaluation of gnns under heterophily: are we really making progress?
\newblock In \emph{International Conference on Learning Representations}, 2023.

\bibitem[Rong et~al.(2020)Rong, Huang, Xu, and Huang]{dropedge}
Rong, Y., Huang, W., Xu, T., and Huang, J.
\newblock Dropedge: Towards deep graph convolutional networks on node classification.
\newblock In \emph{International Conference on Learning Representations}, 2020.

\bibitem[Shlomi et~al.(2021)Shlomi, Battaglia, and Vlimant]{Shlomi2021physics}
Shlomi, J., Battaglia, P., and Vlimant, J.-R.
\newblock Graph neural networks in particle physics.
\newblock 2021.

\bibitem[Stretcu et~al.(2019)Stretcu, Viswanathan, Movshovitz-Attias, Platanios, Ravi, and Tomkins]{stretcu2019agree}
Stretcu, O., Viswanathan, K., Movshovitz-Attias, D., Platanios, E., Ravi, S., and Tomkins, A.
\newblock Graph agreement models for semi-supervised learning.
\newblock In \emph{Advances in Neural Information Processing Systems}, 2019.

\bibitem[Veličković et~al.(2018)Veličković, Cucurull, Casanova, Romero, Liò, and Bengio]{gatv1}
Veličković, P., Cucurull, G., Casanova, A., Romero, A., Liò, P., and Bengio, Y.
\newblock Graph attention networks.
\newblock In \emph{International Conference on Learning Representations}, 2018.

\bibitem[Wang et~al.(2019)Wang, Ying, Huang, and Leskovec]{wang2019improving}
Wang, G., Ying, R., Huang, J., and Leskovec, J.
\newblock Improving graph attention networks with large margin-based constraints.
\newblock In \emph{Graph Representation Learning Workshop, Neural Information Processing Systems}, 2019.

\bibitem[Wu et~al.(2019)Wu, Pan, Long, Jiang, and Zhang]{Wu2019Wavenet}
Wu, Z., Pan, S., Long, G., Jiang, J., and Zhang, C.
\newblock Graph wavenet for deep spatial-temporal graph modeling.
\newblock In \emph{International Joint Conference on Artificial Intelligence}, 2019.

\bibitem[Xu et~al.(2018)Xu, Li, Tian, Sonobe, Kawarabayashi, and Jegelka]{jknets}
Xu, K., Li, C., Tian, Y., Sonobe, T., Kawarabayashi, K.-i., and Jegelka, S.
\newblock Representation learning on graphs with jumping knowledge networks.
\newblock In \emph{International Conference on Machine Learning}, 2018.

\bibitem[Xu et~al.(2019)Xu, Hu, Leskovec, and Jegelka]{xu2018Powerful}
Xu, K., Hu, W., Leskovec, J., and Jegelka, S.
\newblock How powerful are graph neural networks?
\newblock In \emph{International Conference on Learning Representations}, 2019.

\bibitem[Yan et~al.(2022)Yan, Hashemi, Swersky, Yang, and Koutra]{yan2022sides}
Yan, Y., Hashemi, M., Swersky, K., Yang, Y., and Koutra, D.
\newblock Two sides of the same coin: Heterophily and oversmoothing in graph convolutional neural networks.
\newblock In \emph{IEEE International Conference on Data Mining}, 2022.

\bibitem[Yang et~al.(202)Yang, Ma, and Cheng]{propreg}
Yang, H., Ma, K., and Cheng, J.
\newblock Rethinking graph regularization for graph neural networks.
\newblock In \emph{Advances in Neural Information Processing Systems}, 202.

\bibitem[Yang et~al.(2019)Yang, Kang, Cao, Jin, Yang, and Guo]{yang2019top}
Yang, L., Kang, Z., Cao, X., Jin, D., Yang, B., and Guo, Y.
\newblock Topology optimization based graph convolutional network.
\newblock In \emph{International Joint Conference on Artificial Intelligence}, 2019.

\bibitem[Zhang \& Chen(2020)Zhang and Chen]{Zhang2020Inductive}
Zhang, M. and Chen, Y.
\newblock Inductive matrix completion based on graph neural networks.
\newblock In \emph{International Conference on Learning Representations}, 2020.

\bibitem[Zhang et~al.(2021)Zhang, Yang, Sheng, Li, Ouyang, Tao, Yang, and Cui]{ndls}
Zhang, W., Yang, M., Sheng, Z., Li, Y., Ouyang, W., Tao, Y., Yang, Z., and Cui, B.
\newblock Node dependent local smoothing for scalable graph learning.
\newblock In \emph{Advances in Neural Information Processing Systems}, 2021.

\bibitem[Zhao \& Akoglu(2020)Zhao and Akoglu]{pairnorm}
Zhao, L. and Akoglu, L.
\newblock Pairnorm: Tackling oversmoothing in gnns.
\newblock In \emph{International Conference on Learning Representations}, 2020.

\bibitem[Zhou et~al.(2020)Zhou, Huang, Li, Zha, Chen, and Hu]{groupnorm}
Zhou, K., Huang, X., Li, Y., Zha, D., Chen, R., and Hu, X.
\newblock Towards deeper graph neural networks with differentiable group normalization.
\newblock In \emph{Advances in Neural Information Processing Systems}, 2020.

\bibitem[Zhou et~al.(2021)Zhou, Dong, Wang, Lee, Hooi, Xu, and Feng]{nodenorm}
Zhou, K., Dong, Y., Wang, K., Lee, W.~S., Hooi, B., Xu, H., and Feng, J.
\newblock Understanding and resolving performance degradation in graph convolutional networks.
\newblock In \emph{Conference on Information and Knowledge Management}, 2021.

\bibitem[Zou et~al.(2019)Zou, Hu, Wang, Jiang, Sun, and Gu]{ladies}
Zou, D., Hu, Z., Wang, Y., Jiang, S., Sun, Y., and Gu, Q.
\newblock Layer-dependent importance sampling for training deep and large graph convolutional networks.
\newblock In \emph{Advances in Neural Information Processing Systems}, 2019.

\end{thebibliography}
\bibliographystyle{icml2024}

%%%%%%%%%%%%%%%%%%%%%%%%%%%%%%%%%%%%%%%%%%%%%%%%%%%%%%%%%%%%%%%%%%%%%%%%%%%%%%%
%%%%%%%%%%%%%%%%%%%%%%%%%%%%%%%%%%%%%%%%%%%%%%%%%%%%%%%%%%%%%%%%%%%%%%%%%%%%%%%
% APPENDIX
%%%%%%%%%%%%%%%%%%%%%%%%%%%%%%%%%%%%%%%%%%%%%%%%%%%%%%%%%%%%%%%%%%%%%%%%%%%%%%%
%%%%%%%%%%%%%%%%%%%%%%%%%%%%%%%%%%%%%%%%%%%%%%%%%%%%%%%%%%%%%%%%%%%%%%%%%%%%%%%

\newpage
\appendix
\onecolumn
\clearpage
\section{Theoretical Derivations}

\subsection{Derivation of Insight~\ref{insight:GAT}}\label{insightDer:GAT}
\begin{theorem*}[Restated Insight~\ref{insight:GAT}]
 GATs are challenged to switch off neighborhood aggregation during training, as this would require the model to enter a less trainable regime with large attention parameters $\norm{a}^2 >> 1$.   
\end{theorem*}
We have to distinguish GATs with and without weight sharing in our analysis.

\textbf{GATs with weight sharing:}

To investigate the ability of a GAT to switch off neighborhood aggregation, let us focus on a link $(i,j)$ that should neither contribute to the feature transformation of $i$ nor $j$.

This implies that we need to find attention parameters $\mathbf{a}$ (and potentially feature transformations $W$) so that $\alpha_{ij}/\alpha_{ii} << 1$ with $\alpha_{ij}/\alpha_{ii} = \exp\left(e_{ij} - e_{ii}\right)$.
This implies that we require $e_{ij} - e_{ii} << 0$ and thus $\mathbf{a}^T\phi\left(\textbf{W} \left(\textbf{h}_i + \textbf{h}_j\right)\right) - 2 \mathbf{a}^T\phi\left(\textbf{W} \left(\textbf{h}_i \right)\right) << 0$.

Since we also require $\alpha_{ij}/\alpha_{jj} << 1$, it follows from adding both inequalities that $\mathbf{a}^T\left[\phi\left(\textbf{W} \left(\textbf{h}_i + \textbf{h}_j\right)\right) - \left( \phi\left(\textbf{W} \textbf{h}_i \right)+ \phi\left(\textbf{W} \textbf{h}_j \right)\right) \right] << 0$.

This inequality can only be fulfilled if there exists at least one feature $f$ for which 
\begin{align*}%\label{eq:deltaij}
\Delta_{fij} ;= a[f] \left[\phi\left(\textbf{W}[f,:] \left(\textbf{h}_i + \textbf{h}_j\right)\right) - \left( \phi\left(\textbf{W}[f,:] \textbf{h}_i \right)+ \phi\left(\textbf{W} [f,:]\textbf{h}_i \right)\right) \right]
\end{align*}
fulfills $\Delta_{fij}  < < 0$. 
Yet, note that if both $\phi\left(\textbf{W} [f,:]\textbf{h}_i \right)$ and $\phi\left(\textbf{W} [f,:]\textbf{h}_j \right)$ are positive or both are negative, we just get $\Delta_{fij} = 0$ because of the definition of a LeakyReLU.
Thus, there must exist at least one feature $f$ so that without loss of generality $\phi\left(\textbf{W} [f,:]\textbf{h}_i \right) < 0$ and $\phi\left(\textbf{W} [f,:]\textbf{h}_j \right) > 0$. 
% Furthermore, the positive $\phi\left(\textbf{W} [f,:]\textbf{h}_j \right) >> \phi\left(\textbf{W}[f,:] \left(\textbf{h}_i + \textbf{h}_j\right)\right) - \phi\left(\textbf{W}[f,:] \textbf{h}_i \right)  > \phi\left(\textbf{W}[f,:] \left(\textbf{h}_i + \textbf{h}_j\right)\right)$.
%Yet, also 

It follows that if $a[f] > 0$ that
\begin{align*}
0 & > a[f] \phi\left(\textbf{W} [f,:]\textbf{h}_i \right) >> a[f] \left(\phi\left(\textbf{W}[f,:] \left(\textbf{h}_i + \textbf{h}_j\right)\right) - \phi\left(\textbf{W}[f,:] \textbf{h}_j \right) \right)\\
& > a[f] \left(\phi\left(\textbf{W}[f,:] \left(\textbf{h}_i + \textbf{h}_j\right)\right) - 2\phi\left(\textbf{W}[f,:] \textbf{h}_j \right)\right) 
\end{align*}
also receives a negative contribution that makes $\alpha_{ij}/\alpha_{jj}$ smaller.
Yet, what happens to $\alpha_{ij}/\alpha_{ii}$?
By distinguishing two cases, namely $\textbf{W}[f,:] \left(\textbf{h}_i + \textbf{h}_j\right) > 0$ or $\textbf{W}[f,:] \left(\textbf{h}_i + \textbf{h}_j\right) < 0$ and computing 
\begin{align*}
 a[f] \left[\phi\left(\mathbf{W}\left(\textbf{h}_i + \textbf{h}_j\right)\right) - 2\phi\left(\textbf{W}[f,:] \textbf{h}_j \right)\right] > 0
\end{align*}
we find the feature contribution to be positive.

If $a[f] < 0$, then 
\begin{align*}
0 & > a[f] \phi\left(\textbf{W} [f,:]\textbf{h}_j \right) >> a[f] \left(\phi\left(\textbf{W}[f,:] \left(\textbf{h}_i + \textbf{h}_j\right)\right) - \phi\left(\textbf{W}[f,:] \textbf{h}_i \right) \right)\\
& > a[f] \left(\phi\left(\textbf{W}[f,:] \left(\textbf{h}_i + \textbf{h}_j\right)\right) - 2\phi\left(\textbf{W}[f,:] \textbf{h}_i \right)\right) 
\end{align*}
and $\alpha_{ij}/\alpha_{jj}$ is reduced.
Similarly, we can derive that at the same time $\alpha_{ij}/\alpha_{ii}$ is increased, however.

This implies that any feature that contributes to reducing $\Delta_{fij}$ automatically increases one feature while it increases another.
We therefore need multiple features $f$ to contribute to reducing either $\alpha_{ij}/\alpha_{ii}$ or $\alpha_{ij}/\alpha_{jj}$ to compensate for other increases.

This implies, in order to switch off neighborhood aggregation, we would need a high dimensional space of features that cater to switching off specific links without strengthening others. Furthermore, they would need large absolute values of $a[f]$ and norms of $\mathbf{W}[f,:]$ or exploding feature vectors $\mathbf{h}$ to achieve this.

Yet, all these norms are constrained by the derived conservation law and therefore prevent learning a representation that switches off full neighborhoods.
% Returning to the inequalities for the original $\alpha_{ij}$, we see that 
% \begin{align*}%\label{eq:deltaij}
% a[f] \left[\phi\left(\textbf{W}[f,:] \left(\textbf{h}_i + \textbf{h}_j\right)\right) - 2 \phi\left(\textbf{W}[f,:] \textbf{h}_i \right) \right ]
% \end{align*}

\textbf{GATs without weight sharing:}

The flow of argumentation without weight sharing is very similar to the one above with weight sharing.
Yet, we have to distinguish more cases.

Similarly to before, we require $\alpha_{ij}/\alpha_{jj} << 1$ and $\alpha_{ji}/\alpha_{ii} << 1$. 
It follows from adding both related inequalities that 

$$\mathbf{a}^T\left[\phi\left(\textbf{W}_s \textbf{h}_i + \textbf{W}_t \textbf{h}_j\right) + \phi\left(\textbf{W}_s \textbf{h}_j + \textbf{W}_t \textbf{h}_i\right) -  \phi\left(\left(\textbf{W}_s +  \textbf{W}_t\right) \textbf{h}_i \right)- \phi\left(\left(\textbf{W}_s +  \textbf{W}_t\right)  \textbf{h}_j \right) \right] << 0.$$

This implies that for at least one feature $f$, we require
\begin{align}\label{eq:allneg}
\begin{split}
 a[f] & \large[\phi\left(\textbf{W}_s[f,:] \textbf{h}_i + \textbf{W}_t[f,:] \textbf{h}_j\right) + \phi\left(\textbf{W}_s[f,:] \textbf{h}_j + \textbf{W}_t [f,:]\textbf{h}_i\right) \\
 & -  \phi\left(\left(\textbf{W}_s [f,:]+  \textbf{W}_t[f,:]\right) \textbf{h}_i \right)- \phi\left(\left(\textbf{W}_s[f,:] +  \textbf{W}_t[f,:]\right)  \textbf{h}_j \right) \large] << 0. 
 \end{split}
 \end{align}
Again, our goal is to show that this feature automatically decreases the contribution of one feature while it increases another.
As argued above, switching off neighborhood aggregation would therefore need a high dimensional space of features that cater to switching off specific links without strengthening others. Furthermore, they would need large absolute values of $a[f]$ and norms of $\mathbf{W}[f,:]$ or exploding feature vectors $\mathbf{h}$ to achieve this.
Our derived norm constraints, however, prevent learning such a model representation.

Concretely, without loss of generality, we therefore have to show that if 
\begin{align}\label{eq:nb1}
    a[f] \large[\phi\left(\textbf{W}_s[f,:] \textbf{h}_i + \textbf{W}_t[f,:] \textbf{h}_j\right) - \phi\left(\left(\textbf{W}_s[f,:] +  \textbf{W}_t[f,:]\right) \textbf{h}_j \right) < 0,
\end{align}
at the same time, we receive
\begin{align}\label{eq:nb2}
    a[f] \large[\phi\left(\textbf{W}_s[f,:] \textbf{h}_j + \textbf{W}_t[f,:] \textbf{h}_i\right) - \phi\left(\left(\textbf{W}_s[f,:] +  \textbf{W}_t[f,:]\right) \textbf{h}_i \right) > 0,
\end{align}
(or vice versa).

In principle, we have to show this for $16$ different cases of pre-activation sign configurations for the four terms in Eq.~(\ref{eq:allneg}).
Yet, since the argument is symmetric with respect to exchanging $i$ and $j$, only $8$ different cases remain.
Two trivial cases are identical signs for all four terms. 
These are excluded, as the left hand side (LHS) of Eq.~(\ref{eq:allneg}) would become zero and thus not contribute to our goal to switch off neighborhood aggregation.
%One of the remaining six cases is symmetric with respect to flipping positive and negatives signs, which only leaves the following five cases to discuss.
In the following, we will discuss the remaining six cases.
Please note that for the remainder of this derivation $\alpha > 0$ denotes the slope of the leakyReLU and not the attention weights $\alpha_{ij}$.

\begin{enumerate}
\item \textbf{Case $(+-++)$:} Let us assume that $\textbf{W}_s[f,:] \textbf{h}_i + \textbf{W}_t[f,:] \textbf{h}_j > 0$,  $\textbf{W}_s[f,:] \textbf{h}_j + \textbf{W}_t[f,:] \textbf{h}_i < 0$, $\left(\textbf{W}_s[f,:] +  \textbf{W}_t[f,:]\right) \textbf{h}_i > 0$, and $\left(\textbf{W}_s[f,:] +  \textbf{W}_t[f,:]\right) \textbf{h}_j > 0$.

From this assumption and the fact that $\phi$ is a leakyReLU it follows that the LHS of Eq.~(\ref{eq:allneg}) becomes:
$a[f] \large[\phi\left(\textbf{W}_s[f,:] \textbf{h}_i + \textbf{W}_t[f,:] \textbf{h}_j\right) + \phi\left(\textbf{W}_s[f,:] \textbf{h}_j + \textbf{W}_t [f,:]\textbf{h}_i\right) 
  -  \phi\left(\left(\textbf{W}_s [f,:]+  \textbf{W}_t[f,:]\right) \textbf{h}_i \right)- \phi\left(\left(\textbf{W}_s[f,:] +  \textbf{W}_t[f,:]\right)  \textbf{h}_j \right) \large] = a[f] (\alpha-1)\large[ \textbf{W}_s[f,:] \textbf{h}_j + \textbf{W}_t[f,:] \textbf{h}_i\large]$.
 Since $\alpha-1 < 0$ and $\large[ \textbf{W}_s[f,:] \textbf{h}_j + \textbf{W}_t[f,:] \textbf{h}_i\large] < 0$ according to our assumption, Eq.~(\ref{eq:allneg}) demands $a[f] < 0$. 
To switch off neighborhood aggregation, we would need to be able to make the LHS of Eq.~(\ref{eq:nb1}) and Eq.~(\ref{eq:nb2})  Eq.~(\ref{eq:nb2}) negative.
Yet, a negative $a[f]$ leads to a positive LHS of Eq.~(\ref{eq:nb2}).
Thus, the assumed sign configuration cannot support switching off neighborhood aggregation.

\item \textbf{Case $(+---)$:} Let us assume that $\textbf{W}_s[f,:] \textbf{h}_i + \textbf{W}_t[f,:] \textbf{h}_j > 0$,  $\textbf{W}_s[f,:] \textbf{h}_j + \textbf{W}_t[f,:] \textbf{h}_i < 0$, $\left(\textbf{W}_s[f,:] +  \textbf{W}_t[f,:]\right) \textbf{h}_i < 0$, and $\left(\textbf{W}_s[f,:] +  \textbf{W}_t[f,:]\right) \textbf{h}_j < 0$.

The LHS of Eq.~(\ref{eq:allneg}) becomes $a[f] (1-\alpha)\large[ \textbf{W}_s[f,:] \textbf{h}_i + \textbf{W}_t[f,:] \textbf{h}_j\large]$, which demands $a[f] < 0$.
Accordingly, the LHS of Eq.~(\ref{eq:nb1}) is clearly negative, while the LHS of Eq.~(\ref{eq:nb2}) is $a[f] \alpha \textbf{W}_s[f,:] (\textbf{h}_j - \textbf{h}_i) > 0$.
The last inequality follows from our assumptions that imply $\textbf{W}_s[f,:] \textbf{h}_j < \textbf{W}_s[f,:]\textbf{h}_i$ by combining the assumptions $\left(\textbf{W}_s[f,:] +  \textbf{W}_t[f,:]\right) \textbf{h}_j < 0$ and  $\textbf{W}_s[f,:] \textbf{h}_i + \textbf{W}_t[f,:] \textbf{h}_j > 0$.
Again, this result implies that the considered sign configuration does not support switching off neighborhood aggregation.

\item \textbf{Case $(+++-)$:} Let us assume that $\textbf{W}_s[f,:] \textbf{h}_i + \textbf{W}_t[f,:] \textbf{h}_j > 0$,  $\textbf{W}_s[f,:] \textbf{h}_j + \textbf{W}_t[f,:] \textbf{h}_i > 0$, $\left(\textbf{W}_s[f,:] +  \textbf{W}_t[f,:]\right) \textbf{h}_i > 0$, and $\left(\textbf{W}_s[f,:] +  \textbf{W}_t[f,:]\right) \textbf{h}_j < 0$.

The LHS of Eq.~(\ref{eq:allneg}) becomes $a[f] (1-\alpha)\large[ \textbf{W}_s[f,:] \textbf{h}_j + \textbf{W}_t[f,:] \textbf{h}_j\large]$, which demands $a[f] > 0$.
Accordingly, the LHS of Eq.~(\ref{eq:nb1}) becomes positive, which hampers switching-off neighborhood aggregation as discussed.

\item \textbf{Case $(---+)$:} Let us assume that $\textbf{W}_s[f,:] \textbf{h}_i + \textbf{W}_t[f,:] \textbf{h}_j < 0$,  $\textbf{W}_s[f,:] \textbf{h}_j + \textbf{W}_t[f,:] \textbf{h}_i < 0$, $\left(\textbf{W}_s[f,:] + \textbf{W}_t[f,:]\right) \textbf{h}_i < 0$, and $\left(\textbf{W}_s[f,:] +  \textbf{W}_t[f,:]\right) \textbf{h}_j > 0$.

The LHS of Eq.~(\ref{eq:allneg}) becomes $a[f] (\alpha-1)\large[ \textbf{W}_s[f,:] \textbf{h}_j + \textbf{W}_t[f,:] \textbf{h}_j\large]$, which demands $a[f] > 0$.
Accordingly, the LHS of Eq.~(\ref{eq:nb1}) becomes clearly negative. 
However, the LHS of Eq.~(\ref{eq:nb2}) is positive, as $a[f] \alpha \textbf{W}_s[f,:] (\textbf{h}_j - \textbf{h}_i) > 0$.

The last inequality follows from our assumptions that imply $\textbf{W}_s[f,:] \textbf{h}_j > \textbf{W}_s[f,:]\textbf{h}_i$ by combining the assumptions $\left(\textbf{W}_s[f,:] +  \textbf{W}_t[f,:]\right) \textbf{h}_j > 0$ and  $\textbf{W}_s[f,:] \textbf{h}_i + \textbf{W}_t[f,:] \textbf{h}_j < 0$.
Again, this analysis implies that the considered sign configuration does not support switching off neighborhood aggregation.

\item \textbf{Case $(+-+-)$:} Let us assume that $\textbf{W}_s[f,:] \textbf{h}_i + \textbf{W}_t[f,:] \textbf{h}_j > 0$,  $\textbf{W}_s[f,:] \textbf{h}_j + \textbf{W}_t[f,:] \textbf{h}_i < 0$, $\left(\textbf{W}_s[f,:] +  \textbf{W}_t[f,:]\right) \textbf{h}_i > 0$, and $\left(\textbf{W}_s[f,:] +  \textbf{W}_t[f,:]\right) \textbf{h}_j < 0$.

According to our assumptions the LHS of Eq.~(\ref{eq:nb1}) can only be negative if $a[f] < 0$.
Yet, the LHS of Eq.~(\ref{eq:nb2}) can only be negative if $a[f] > 0$. 
Thus, this case clearly cannot contribute to switching off neighborhood aggregation.
%Eq.~(\ref{eq:allneg}) becomes $a[f] (1-\alpha) \textbf{W}_t[f,:] \left(\textbf{h}_j - \textbf{h}_i\right) < 0$.  
%which demands $a[f] < 0$.
% Accordingly, the LHS of Eq.~(\ref{eq:nb2}) 
%While this does not directly determine the sign of $a[f]$, Eq.~(\ref{eq:nb2}) gives us clearer information.
%Note that according to our assumptions the LHS of Eq.~(\ref{eq:nb2}) can only be negative if $a[f] > 0$.

%This implies that the LHS of Eq.~(\ref{eq:nb1}) is $a[f] \large[ \textbf{W}_s[f,:] \textbf{h}_i + ((1-\alpha)\textbf{W}_t[f,:] -\alpha \textbf{W}_s[f,:])\textbf{h}_j \large] > 0$.
%We can see this by 

%The LHS of Eq.~(\ref{eq:nb2}) is $a[f] \large[ \alpha\textbf{W}_s[f,:] \textbf{h}_j - (\alpha \textbf{W}_s[f,:] + (1-\alpha) \textbf{W}_t[f,:])\textbf{h}_i \large]$.
% Accordingly, 

\item \textbf{Case $(+--+)$:} Let us assume that $\textbf{W}_s[f,:] \textbf{h}_i + \textbf{W}_t[f,:] \textbf{h}_j > 0$,  $\textbf{W}_s[f,:] \textbf{h}_j + \textbf{W}_t[f,:] \textbf{h}_i < 0$, $\left(\textbf{W}_s[f,:] +  \textbf{W}_t[f,:]\right) \textbf{h}_i < 0$, and $\left(\textbf{W}_s[f,:] +  \textbf{W}_t[f,:]\right) \textbf{h}_j > 0$.

Eq.~(\ref{eq:allneg}) becomes $a[f] (1-\alpha) \textbf{W}_s[f,:] \left(\textbf{h}_i - \textbf{h}_j\right) < 0$.  
At the same time, the LHS of Eq.~(\ref{eq:nb1}) simplifies to $a[f] \textbf{W}_s[f,:] (\textbf{h}_i - \textbf{h}_j)$ and the LHS of Eq.~(\ref{eq:nb2}) is $a[f] \alpha \textbf{W}_s[f,:] (\textbf{h}_j - \textbf{h}_i) > 0$.

Hence, a negative Eq.~(\ref{eq:allneg}) leads to a positive Eq.~(\ref{eq:nb2}).
Accordingly, the last possible sign configuration also does not support switching off neighborhood aggregation, which concludes our derivation. 
\end{enumerate}

\subsection{Proof of Theorem~\ref{theoremStructureGATE}}
\begin{theorem*}[Restated Theorem \ref{theoremStructureGATE}]
The gradients and parameters of GATE for layer $l \in [L-1]$ are conserved according to the following laws:
\begin{equation}
\dotP{W^l[i,:]}{\nabla_{W^l[i,:]}\mathcal{L}}  = \dotP{W^{l+1}[:,i]}{\nabla_{W^{l+1}[:,i]}\mathcal{L}} + \dotP{a^{l+1}_s[i]}{\nabla_{a^{l+1}_s[i]} \mathcal{L}} + \dotP{a^{l+1}_t[i]}{\nabla_{a^{l+1}_t[i]} \mathcal{L}} .
\label{eqStructureOfGradsGATE}
\end{equation}
and, if additional independent matrices $\mathbf{U}^l$ and $\mathbf{V}^l$ are trainable, it also holds 
\begin{equation}
\dotP{a^l_s[i]}{\nabla_{a^l_s[i]} \mathcal{L}} + \dotP{a^l_t[i]}{\nabla_{a^l_t[i]} \mathcal{L}}  = \dotP{U^{l}[i,:]}{\nabla_{U^{l}[i,:]}\mathcal{L}} + \dotP{V^{l}[i,:]}{\nabla_{V^{l}[i,:]}\mathcal{L}}. 
\label{eqStructureOfGradsGATE2}
\end{equation}
\end{theorem*}

The proof is analogous to the derivation of Theorem 2.2 by \citep{balanceGATs} that is restated in this work as Theorem \ref{theoremStructureOfGrads}. For ease, we replicate their notation and definitions here. 

\begin{theorem*}[Rescale invariance: Def 5.1 by \citet{balanceGATs}]
The loss $\mathcal{L}(\theta)$ is rescale-invariant with respect to disjoint subsets of the parameters $\theta_1$ and $\theta_2$ if for every $\lambda > 0$ we have $\mathcal{L}(\theta) = \mathcal{L}( (\lambda \theta_1, \lambda^{-1}\theta_2, \theta_d))$, 
 where $\theta = (\theta_1, \theta_2, \theta_d)$.
\end{theorem*}

\begin{theorem*}[Gradient structure due to rescale invariance Lemma 5.2 in \cite{balanceGATs}]
The rescale invariance of $\mathcal{L}$ enforces the following geometric constraint on the gradients of the loss with respect to its parameters:
\begin{equation}\label{eqGradStruct}
\dotP{ \theta_1}{\nabla_{\theta_1}\mathcal{L}} -  \dotP{\theta_2}{ \nabla_{\theta_2}\mathcal{L}} = 0.
\end{equation}
\label{gradientStructure}
\end{theorem*}

We first consider the simpler case of GATE$_S$, i.e. $W=U=V$ 

\begin{theorem}[Structure of GATE$_S$ gradients]
The gradients and parameters of GATE$_S$ for layer $l \in [L-1]$ are conserved according to the following laws:
\begin{equation}
\dotP{W^l[i,:]}{\nabla_{W^l[i,:]}\mathcal{L}}  = \dotP{W^{l+1}[:,i]}{\nabla_{W^{l+1}[:,i]}\mathcal{L}} + \dotP{a^{l}_s[i]}{\nabla_{a^{l}_s[i]} \mathcal{L}} + \dotP{a^{l}_t[i]}{\nabla_{a^{l}_t[i]} \mathcal{L}} .
\label{eqStructureOfGradsGATES}
\end{equation}
\end{theorem}

Following a similar strategy to \citep{balanceGATs}, we identify rescale invariances for every neuron $i$ at layer $l$ that induce the stated gradient structure.

Given the following definition of disjoint subsets $\theta_1$ and $\theta_2$ of the parameter set $\theta$, associated with neuron $i$ in layer $l$,

\begin{align*}
 \theta_1&=\{x \vert x \in W^l[i,:]\} \\ %\cup \{x \vert x \in U^l[i,:]\} \cup \{x \vert x \in V^l[i,:]\} \\
 \theta_2&=\{w \vert w \in W^{l+1}[:,i]\} \cup \{a^l_s[i]\}  \cup \{a^l_t[i]\}
\end{align*}

We show that the loss of GATE$_S$ remains invariant for any $\lambda > 0$. 

The only components of the network that potentially change under rescaling are $h_u^{l}[i]$, $h_v^{l+1}[j]$, and $\alpha^l_{uv}$.

The scaled network parameters are denoted with a tilde as $\tilde{a_s}^l[i] = \lambda^{-1} a_s^l[i]$, $\tilde{a_t}^l[i] = \lambda^{-1} a_t^l[i]$, and $\tilde{W}^l[i,j] = \lambda W^l[i,j]$, and the corresponding networks components scaled as a result are denoted by $\tilde{h}_u^{l}[i]$, $\tilde{h}_v^{l+1}[k]$, and $\tilde{\alpha}^l_{uv}$.

We show that the parameters of upper layers remain unaffected, as $\tilde{h}_v^{l+1}[k]$ coincides with its original non-scaled variant $\tilde{h}_v^{l+1}[k] = h_v^{l+1}[k]$. 

Also recall Eq. (\ref{GATEdef}) for $W=U=V$ as: 
$$e_{uv}^l = (( 1 - q_{uv} )a_s^l + (q_{uv} ) a_t^l) ^\top \cdot \phi (W^l h_u^{l-1} + W^l h_v^{l-1}) $$ 
where $q_{uv}=1$ if $u=v$ and $q_{uv}=0$ if $u\neq v$. 

For simplicity, we rewrite this as: 
\begin{align}
e_{uv,u\neq v}^l &= (a_s^l)^\top \cdot \phi (W^l h_u^{l-1} + W^l h_v^{l-1})\\
e_{uv,u=v}^l &= (a_t^l)^\top \cdot \phi (W^l h_u^{l-1} + W^l h_v^{l-1})
\end{align}

We show that
      
\begin{align}
\tilde{\alpha}_{uv}^l &= \frac{\text{exp}(\tilde{e}^l_{uv})}{\sum_{u'\in\mathcal{N}(v)} \text{exp}( \tilde{e}^l_{uv})} =  \alpha_{uv}^l \;,\;\; \text{because}\\
    \tilde{e}_{uv,u\neq v}^l &= e_{uv,u\neq v}^l \;,\;\; \text{and} \;\; 
    \tilde{e}_{uv,u=v}^l = e_{uv,u=v}^l 
\end{align}

 which follows from the positive homogeneity of $\phi$ that allows
 
 \begin{align}
   \tilde{e}_{uv,u=v}^l &=  \lambda^{-1} a_s^l[i] \phi (\sum_j^{n_{l-1}} \lambda W^l[i,j] (h_u^{l-1}[j] + h_v^{l-1}[j]) \nonumber \\ &+ \sum_{i'\neq i}^{n_l} a_s^l[i'] \phi (\sum_j^{n_{l-1}} W^l[i',j] (h_u^{l-1}[j] + h_v^{l-1}[j])   \\&= \lambda^{-1} \lambda a_s^l[i] \phi ( \sum_j^{n_{l-1}} W^l[i,j] (h_u^{l-1}[j] + h_v^{l-1}[j]) \nonumber \\ &+ \sum_{i'\neq i}^{n_l} a_s^l[i'] \phi (\sum_j^{n_{l-1}} W^l[i',j] (h_u^{l-1}[j] + h_v^{l-1}[j])   \\ &= e_{uv,u\neq v}^l.
   \end{align}

and similarly, 

\begin{align}
   \tilde{e}_{uv,u=v}^l &=  \lambda^{-1} a_t^l[i] \phi (\sum_j^{n_{l-1}} \lambda W^l[i,j] (h_u^{l-1}[j] + h_v^{l-1}[j]) \nonumber \\ &+ \sum_{i'\neq i}^{n_l} a_t^l[i'] \phi (\sum_j^{n_{l-1}} W^l[i',j] (h_u^{l-1}[j] + h_v^{l-1}[j])   \\&= \lambda^{-1} \lambda a_t^l[i] \phi ( \sum_j^{n_{l-1}} W^l[i,j] (h_u^{l-1}[j] + h_v^{l-1}[j]) \nonumber \\ &+ \sum_{i'\neq i}^{n_l} a_t^l[i'] \phi (\sum_j^{n_{l-1}} W^l[i',j] (h_u^{l-1}[j] + h_v^{l-1}[j])   \\ &= e_{uv,u=v}^l.
   \end{align}

Since $\tilde{\alpha}_{uv}^l = \alpha_{uv}^l$, it follows that 
\begin{align*}
    \tilde{h_u}^{l}[i] &= \phi_1 \left(\sum_{z\in \mathcal{N}(u)} \alpha_{zu}^l \sum_{j}^{n_{l-1}} \lambda W^l[i,j]h_z^{l-1}[j] \right) \\&= \lambda \phi_1 \left(\sum_{z\in \mathcal{N}(u)} \alpha_{zu}^l \sum_{j}^{n_{l-1}} W^l[i,j] h_z^{l-1}[j] \right)\\ &= \lambda h_u^{l}[i].
\end{align*}
In the next layer, we therefore have 
\begin{align*}
\tilde{h}_v^{l+1}[k]  &=  \phi_1 \left(\sum_{u\in \mathcal{N}(v)} \alpha_{uv}^{l+1} \sum_{i}^{n_{l}} \lambda^{-1} W^{l+1}[k,i] \tilde{h}_u^{l}[i] \right) \\ &= \phi_1 \left(\sum_{u\in \mathcal{N}(v)} \alpha_{uv}^{l+1} \sum_{i}^{n_{l}} \lambda^{-1} W^{l+1}[k,i] \lambda h_u^{l}[i] \right) \\ &= \phi_1 \left(\sum_{u\in \mathcal{N}(v)} \alpha_{uv}^{l+1} \sum_{i}^{n_{l}}  W^{l+1}[k,i]  h_u^{l}[i] \right) \\ &=  h_v^{l+1}[k].
\end{align*}

Thus, the output node representations of the network remain unchanged, and the loss $\mathcal{L}$ is rescale-invariant.

Next consider the case that $W^l$, $U^l$, and $V^l$ are independent matrices.
Similarly to the previous reasoning, we see that if we scale $\tilde{W}^l[i,:] = W^l[i,:] \lambda$, then also scaling $\tilde{W}^{l+1}[:,i] = W^{l+1}[:,i] \lambda^{-1}$ and  $\tilde{a}^{l+1}_s[i] = a^{l+1}_s[i] \lambda^{-1}$ and $\tilde{a}_t^{l+1}[i] = a^{l+1}_t[i] \lambda^{-1}$ will keep the GATE layer unaltered.

In this case, we obtain an additional rescaling relationship between $a^l_s$, $a^l_t$ and $U^l$, $V^l$.
A rescaling of the form $\tilde{a_s}^l[i] = \lambda^{-1} a_s^l[i]$, $\tilde{a_t}^l[i] = \lambda^{-1} a_t^l[i]$ could be compensated by $\tilde{U}^l[i,:] = U^l[i,:] \lambda$ and $\tilde{V}^l[i,:] = V^l[i,:] \lambda$.
It follows immediately that $\tilde{e}_{uv} = e_{uv}$.

\subsection{Derivation of Insight~\ref{insight:GATE}}
\label{insightDer:GATE}
Following the analysis in \ref{insightDer:GAT}, in contrast to GAT, $\alpha_{ij}/\alpha_{ii} << 1$ can be easily realized in GATE with $a_s[f] < 0$ and $a_t[f] > 0$ for all or only a subset of the features. Note that for the non-weight-sharing case, $\mathbf{U}$ and $\textbf{V}$ in GATE would simply correspond to $\textbf{W}_s$ and $\textbf{W}_t$, respectively, in GATE and the same line of reasoning holds. Large norms are usually not required to create a notable difference in size between $e_{ii}$ and $e_{ij}$.

\section{Experimental Settings}
\label{expSettings}

Our complete experimental setup is described as follows.

 \paragraph{Non-linearity} For GAT$_S$ and GAT networks, we substitute $\phi$ in Eq. (\ref{GATdef3}) with LeakyReLU as defined in the standard architecture. For GATE, we substitute $\phi$ in Eq. (\ref{GATEdef}) with ReLU in order to be able to interpret the sign of $\mathbf{a_s}$ and $\mathbf{a_t}$ parameters as contributing positively or negatively to neighborhood aggregation. MLP, MLP$_{+GAT}$, and FAGCN also all use ReLU after every hidden layer. 

\paragraph{Network Width} We vary the depth of GAT and GATE networks in all our experiments as specified. For synthetic datasets, the network width is fixed to 64 in all cases. For OGB datasets, we use the hidden dimensions used by \cite{gat}. For the remaining datasets, the network width is also fixed to 64. 

\paragraph{Initialization} The feature transformation parameter matrices, i.e., $\mathbf{W},\mathbf{U}$, and $\mathbf{V}$ are initialized randomly with an orthogonal looks-linear structure \citep{dyniso} for MLP, MLP$_{+GAT}$, GAT$_{(S)}$ and GATE$_{(S)}$.
The parameters $\mathbf{a}$ in GAT$_{(S)}$ use Xavier initialization \citep{GlorotInit}, as is the standard. 
In GATE$_{(S)}$, $\mathbf{a}_s$ and $\mathbf{a}_t$ are initialized to $0$ to initially give equal weights to the features of a node itself and its neighboring nodes.  

\paragraph{Optimization} Synthetic, OGB, and remaining real-world tasks are run for a maximum of $10000$, $2000$, $5000$ epochs, respectively, using the Adam optimizer. To isolate the effect of the architecture and study the parameter dynamics during training as best as possible, we do not use any additional elements such as weight decay and dropout regularization.  We also do not perform any hyperparameter optimization.
However, the learning rate is adjusted for different real-world datasets to enable stable training of models as specified in Table \ref{datasetDetailsAndLR}.
Nevertheless, for a fair comparison, the same learning rate is used for a given problem across all architectures. 
For all synthetic data, a learning rate of $0.005$ is used.  Real-world datasets use their standard train/test/validation splits, i.e. those provided by Pytorch Geometric for Planetoid datasets Cora and Citeseer, by OGB framework for OGB datasets, and by \cite{criticalHet} for all remaining real-world datasets. 

\paragraph{Code} Our experimental code and synthetic data generators are available at \url{https://github.com/RelationalML/GATE.git}.

\begin{table}[h]
\caption{Details of real-world datasets used in experiments.}
\label{datasetDetailsAndLR}
\centering
\begin{tblr}{
  width = \linewidth,
  colspec = {Q[150]Q[110]Q[110]Q[100]Q[115]},
  cells = {c},
  hline{1,15} = {-}{0.08em},
  hline{2} = {-}{0.05em},
}
dataset   & \# nodes & \# edges & \# features & \# classes & learning rate used for $L$ layer networks \\
ogb-arxiv & $169,343$ & $2,315,598$ & $128$ & $40$ & $L=[12]: 0.001$\\
ogb-products & $2,449,029$ & $123,718,152$ & $100$  & $47$ & $L=[8]: 0.001$\\
ogb-mag & $736,389$ & $10,792,672$ & $128$ & $349$ & $L=[5]: 0.005$ & \\
roman-empire & $22,662$ & $32,927$ & $300$& $18$& $L=[5]: 0.001, L=[10]: 0.0005$ \\
amazon-ratings & $24,492$& $93,050$ & $300$& $5$& $L=[5]: 0.001, L=[10]: 0.0005$ \\
questions & $48,921$& $153,540$ & $301$ &  $2$ & $L=[5]: 0.001, L=[10]: 0.0005$ \\
minesweeper & $10,000$ & $39,402$ & $7$ & $2$ & $L=[5]: 0.001, L=[10]: 0.0005$ \\
tolokers & $11,758$ & $519,000$ & $10$ & $2$ & $L=[5]: 0.001, L=[10]: 0.0005$ \\
cora      & $2,708$   & $10,556$  & $1,433$      & $7$        &   $L=[2,5]: 0.005, L=[10,20]: 0.0005$ \\
citeseer  & $3,327$   & $9,104$   & $3,703$      & $6$        & $L=[2,5]: 0.001, L=[10,20]: 0.0001$ \\
actor     & $7,600$   & $26,659$  & $932$       & $5$       & $L=[2,5,10,20]: 0.005$  \\
texas     & $183$    & $279$    & $1,703$      & $5$        &  $L=[2,5]: 0.01, L=[10,20]: 0.0005$\\
wisconsin & $251$    & $450$    & $1,703$      & $5$    &  $L=[2,5]: 0.01, L=[10,20]: 0.005$
\end{tblr}
\end{table}

 \section{Additional Results}
 \label{appendix:additional-results}

\paragraph{Smaller Real-World Datasets} We evaluate GAT and GATE on five small-scale real-world datasets with varying homophily levels $\beta$ as defined in \citep{geomgcn} and report results in Table \ref{table:real-small-data}. Higher values of $\beta$ indicate higher homophily, i.e. similar nodes (with the same label) tend to be connected. We note that a 2-layer network of a baseline method for heterophilic datasets, Geom-GCN \citep{geomgcn}, attains test accuracy (\%) of $64.1$, $67.6$, and $31.6$ for Wisconsin, Texas, and Actor datasets, respectively, which is in line with that achieved by GATE. Except for Citeseer, the best overall performance for each dataset is achieved on a shallow model. This is not surprising as these datasets are small-scale and potentially prone to over-fitting in large models, particularly since we do not use any skip connections or regularization to retain model performance. Furthermore, the three heterophilic datasets have been recently shown to be problematic \cite{criticalHet}. Therefore, a better evaluation of GATE is on relatively large-scale OGB datasets \cite{ogb} and more recent heterophilic datasets \cite{criticalHet} that can exploit the flexibility of GATE. Although GATE is more parameterized than GAT, it usually requires fewer training epochs and generalizes better, in addition to other advantages over GAT as discussed in the paper.

\begin{table}[h]
\caption{Test accuracy ($\%$) of GAT and GATE models for network depth $L$ on small-scale real-world datasets with varying homophily levels $\beta$. Entries marked with * indicate models that achieve $100\%$ training accuracy and stable test accuracy. Otherwise, test accuracy at max. validation accuracy is reported. }
\label{table:real-small-data}
\centering
\begin{tblr}{
  width = \linewidth,
  colspec = {Q[40]Q[40]Q[60]Q[67]Q[67]Q[60]Q[67]Q[67]Q[60]Q[67]Q[67]Q[60]Q[67]Q[67]},
  cells = {c},
  cell{1}{1} = {r=2}{},
  cell{1}{2} = {r=2}{},
  cell{1}{3} = {c=3}{0.201\linewidth},
  cell{1}{6} = {c=3}{0.201\linewidth},
  cell{1}{9} = {c=3}{0.201\linewidth},
  cell{1}{12} = {c=3}{0.201\linewidth},
  hline{1,8} = {-}{0.08em},
  hline{2-3} = {3-14}{0.03em},
}
Data  & $\beta$   & $L=2$   &         &         & $L=5$  &          &         & $L=10$  &         &         & $L=20$  &          &        \\
      &        & GAT     & GATE$_S$ & GATE    & GAT    & GATE$_S$  & GATE    & GAT     & GATE$_S$ & GATE    & GAT     & GATE$_S$  & GATE   \\
Wisc. & $0.21$ & $62.7$* & $\textbf{80.4}$  & $70.5$* & $51.0$ & $\textbf{70.5}$   & $60.7$* & $45.1$  & $\textbf{62.7}$  & $58.8$  & $47.1$  & $\textbf{62.7}$   & $60.7$ \\
Texas & $0.11$ & $56.7$* & $\textbf{67.6}$* & $\textbf{67.6}$* & $51.4$ & $\textbf{67.6}$*  & $\textbf{67.6}$* & $56.7$* & $62.2$* & $\textbf{62.3}$* & $59.4$* & $62.1$*  & $\textbf{64.9}$ \\
Actor & $0.24$ & $27.1$  & $\textbf{32.2}$  & $31.6$  & $25.4$ & $27.5$   & $\textbf{29.2}$  & $25.3$  & $27.4$  & $\textbf{27.9}$  & $24.5$  & $24.6$   & $\textbf{29.4}$ \\
Cora  & $0.83$ & $80.0$  & $\textbf{81.0}$* & $80.8$  & $79.8$ & $\textbf{80.8}$* & $80.4$  & $77.6$  & $\textbf{80.0}$* & $79.2$  & $77.7$  & $77.2$*  & $\textbf{79.0}$ \\
Cite. & $0.71$ & $68.0$  & $67.6$* & $\textbf{68.3}$  & $67.2$ & $\textbf{68.7}$*  & $67.8$  & $66.9$  & $\textbf{67.6}$* & $\textbf{67.6}$  & $68.2$  & $67.1$* & $\textbf{69.2}$ 
\end{tblr}
\end{table}

\begin{figure} [h]
\centering
\includegraphics[width=.75\linewidth]{Figures/alphaDistribution/alpha_ii_distribution_legend.png}
\begin{subfigure}{.125\textwidth}
  \centering
\frame{\includegraphics[width=.97\linewidth]{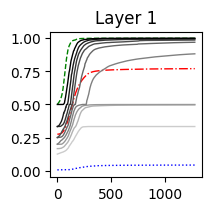}}
\end{subfigure}%
\begin{subfigure}{.25\textwidth}
  \centering  
\frame{\includegraphics[width=.95\linewidth]{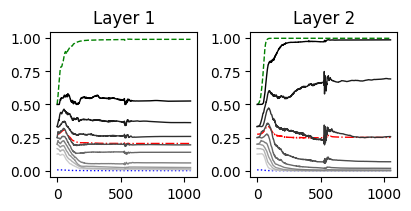}}
\end{subfigure}%
\begin{subfigure}{.625\textwidth}
  \centering  
\frame{\includegraphics[width=.95\linewidth]{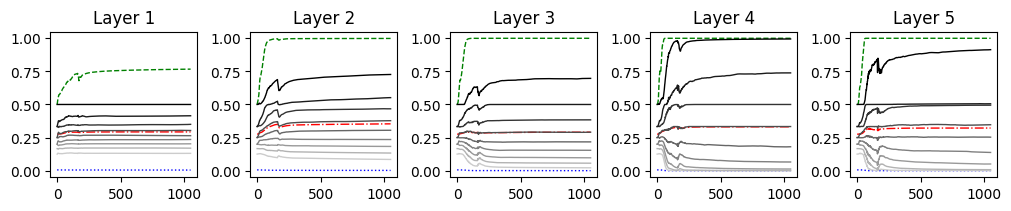}}
\end{subfigure}
  \caption{Distribution of $\alpha_{vv}$ against training epoch for self-sufficient learning problem using the Cora structure with random labels, where input node features are a one-hot encoding of node labels for GAT with attention parameters $\mathbf{a}$ initialized to zero. Left to right: $1$, $2$ and $5$ layer models that achieve test accuracy of $100\%$, $52.7\%$, and $36.2\%$, respectively, which is similar to the results obtained by standard Xavier initialization of attention parameters in GAT.}
  \label{gat-zero-attn}
\end{figure}

\paragraph{Initialization of attention parameters in GAT} We show in Fig. \ref{gat-zero-attn} that setting the initial value of attention parameters $a_s$ and $a_t$ in GATE to zero is, in fact, not what enables neighborhood aggregation but rather the separation of $a$ into $a_s$ and $a_t$ as discussed in Insight \ref{insight:GATE}.

\paragraph{Further analysis of experiments}
We present the analysis of $\alpha$ coefficients learned for some experiments in the main paper that were deferred to the appendix due to space limitations. 

\begin{figure} [h]
\centering
\includegraphics[width=.75\linewidth]{Figures/alphaDistribution/alpha_ii_distribution_legend.png}\\
\begin{subfigure}{.5\textwidth}
  \centering  
\frame{\includegraphics[width=.95\linewidth]{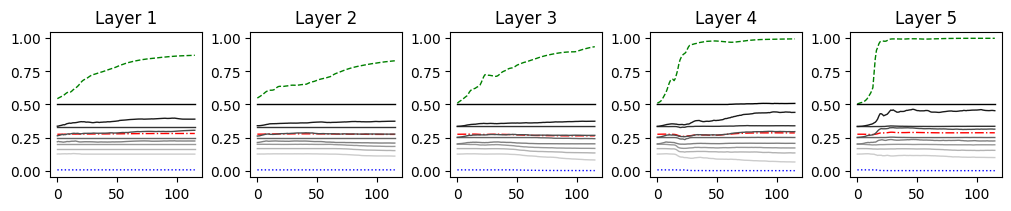}}
\caption{GAT with original labels}
\end{subfigure}%
\begin{subfigure}{.5\textwidth}
  \centering  
\frame{\includegraphics[width=.95\linewidth]{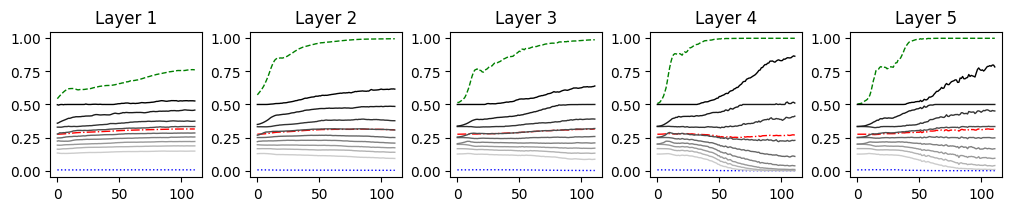}}
\caption{GAT with random labels}
\end{subfigure}\\
\begin{subfigure}{.5\textwidth}
  \centering  
\frame{\includegraphics[width=.95\linewidth]{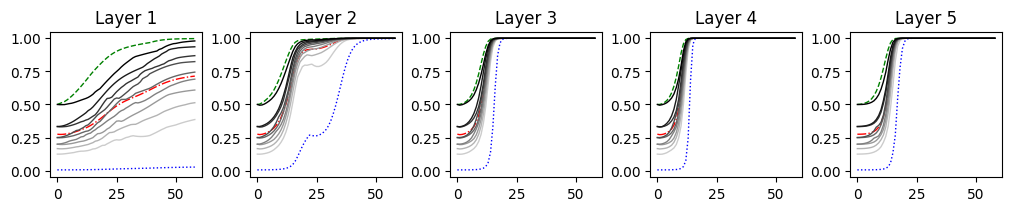}}
\caption{GATE with original labels}
\end{subfigure}%
\begin{subfigure}{.5\textwidth}
  \centering  
\frame{\includegraphics[width=.95\linewidth]{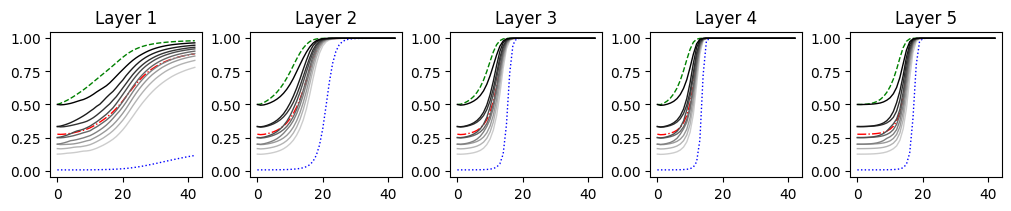}}
\caption{GATE with random labels}
\end{subfigure}  
  \caption{ Distribution of $\alpha_{vv}$ against training epoch for self-sufficient learning problem using Cora structure and input node features as the one-hot encoding of labels for $5$ layer models.}
  \label{alphaDist-cora-oneHotFeats-5-layer}
\end{figure}

\begin{figure} [h]
\centering
\includegraphics[width=.75\linewidth]{Figures/alphaDistribution/alpha_ii_distribution_legend.png}\\
\begin{subfigure}{\textwidth}
\begin{subfigure}{.125\textwidth}
  \centering
\frame{\includegraphics[width=.97\linewidth]{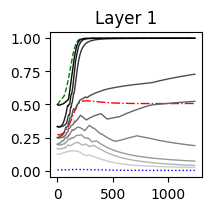}}
\end{subfigure}%
\begin{subfigure}{.25\textwidth}
  \centering  
\frame{\includegraphics[width=.95\linewidth]{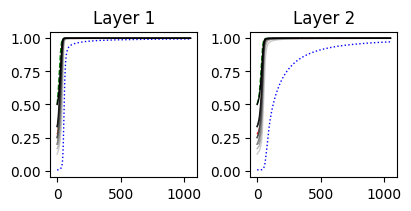}}
\end{subfigure}%
\begin{subfigure}{.625\textwidth}
  \centering  
\frame{\includegraphics[width=.95\linewidth]{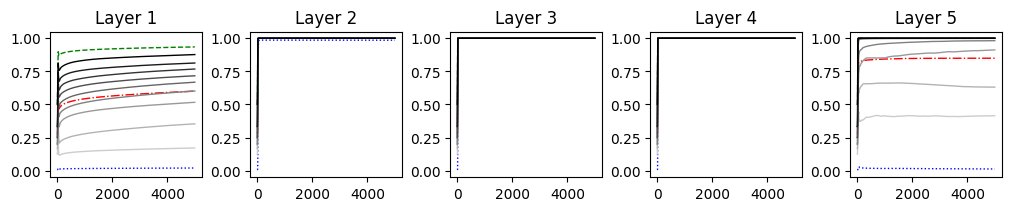}}
\end{subfigure}
\caption{original labels}
\end{subfigure}
\begin{subfigure}{\textwidth}
\begin{subfigure}{.125\textwidth}
  \centering
\frame{\includegraphics[width=.97\linewidth]{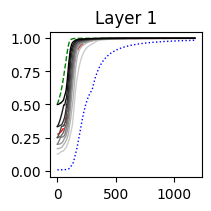}}
\end{subfigure}%
\begin{subfigure}{.25\textwidth}
  \centering  
\frame{\includegraphics[width=.95\linewidth]{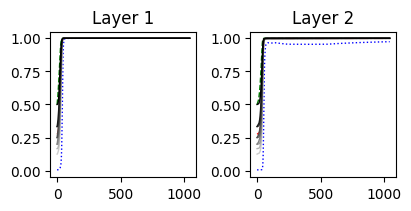}}
\end{subfigure}%
\begin{subfigure}{.625\textwidth}
  \centering  
\frame{\includegraphics[width=.95\linewidth]{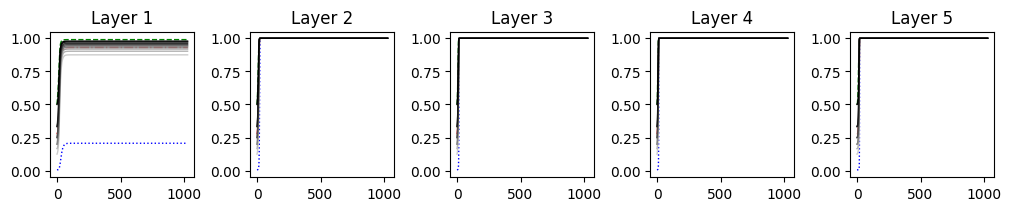}}
\end{subfigure}
\caption{Random labels}
\end{subfigure}
  \caption{ Distribution of $\alpha_{vv}$ against training epoch for the self-sufficient learning problem using Cora graph structure with original (top) and random (bottom) node labels and input node features as a one-hot encoding of labels. Left to right: $1$, $2$, and $5$ layer  GATE$_S$ models that all $100\%$ test accuracy except in the case of 5 layer model using original labels. In this case, although a training accuracy if $100\%$ is achieved at 32 epochs with test accuracy $97.3\%$, a maximum test accuracy of $98.4\%$ is reached at 7257 epochs. Training the model to run to 15000 epochs only increases it to $98.4\%$. An increased learning rate did not improve this case. However, we also run the GAT model for 15000 epochs for this case, and it achieves $85.9\%$ test accuracy at epoch 47 where the model achieves $100\%$ accuracy and only achieves a maximum test accuracy of $89.3\%$ briefly at epoch 8.}
  \label{alphaDist-cora-oneHotFeats-gates}
\end{figure}

\begin{figure} [h]
\centering
\includegraphics[width=.75\linewidth]{Figures/alphaDistribution/alpha_ii_distribution_legend.png}
\begin{subfigure}{.17\textwidth}
  \centering
\frame{\includegraphics[width=.99\linewidth]{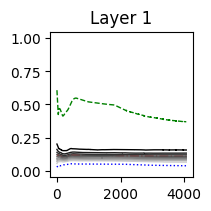}}
\end{subfigure}%
\begin{subfigure}{.33\textwidth}
  \centering  
\frame{\includegraphics[width=.98\linewidth]{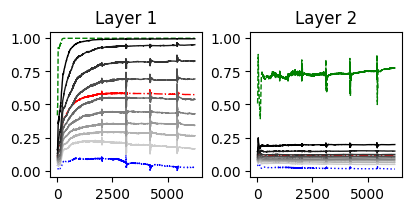}}
\end{subfigure}%
\begin{subfigure}{.5\textwidth}
  \centering  
\frame{\includegraphics[width=.97\linewidth]{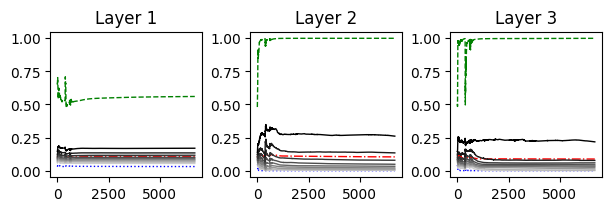}}
\end{subfigure}\\
\begin{subfigure}{.17\textwidth}
  \centering
\frame{\includegraphics[width=.99\linewidth]{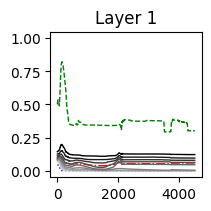}}
\end{subfigure}%
\begin{subfigure}{.33\textwidth}
  \centering  
\frame{\includegraphics[width=.98\linewidth]{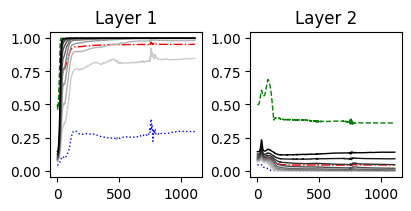}}
\end{subfigure}%
\begin{subfigure}{.5\textwidth}
  \centering  
\frame{\includegraphics[width=.97\linewidth]{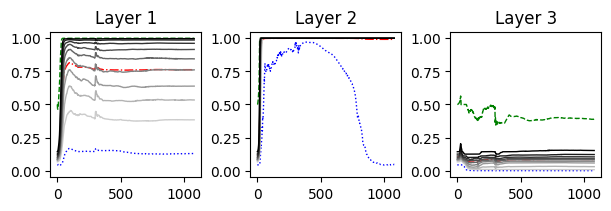}}
\end{subfigure}
  \caption{Distribution of $\alpha_{vv}$ against training epoch for the neighbor-dependent learning problem with $k=1$. Rows: GAT (top) and GATE (bottom) architecture. Columns (left to right): $1$, $2$, and $3$ layer models. While GAT is unable to switch off neighborhood aggregation in any layer, only $1$ layer of the $2$ and $3$ layer models perform neighborhood aggregation.}
  \label{appx:alpha-dist-randSynData-k1}
\end{figure}

\begin{table*}[h]
\centering
\caption{Neighbor-dependent learning: $k$ and $L$ denote the number of aggregation steps of the random GAT used for label generation and the number of layers of the evaluated network, respectively. Entries marked with * identify models where $100\%$ train accuracy is not achieved. Underlined entries identify the model with the highest train accuracy at the epoch of max. test accuracy. This provides an insight into how similar the function represented by the trained model is to the function used to generate node labels, i.e. whether the model is simply overfitting to the train data or really learning the task. Higher training and test accuracy simultaneously indicate better learning. In this regard, the difference in train accuracy at max. test accuracy between GATE and GAT$_S$ or GAT is only $0.4$, $1.0$ and $0.6$ for the settings $(k=1,L=3)$, $(k=2,L=4)$ and $(k=3,L=3)$, respectively. }
\label{table:neighbor-dependent-learning-one-run}

\begin{tblr}{
  width = \linewidth,
  colspec = {Q[20]Q[20]Q[142]Q[142]Q[142]Q[144]Q[146]Q[142]},
  cells = {c},
  cell{1}{1} = {r=2}{},
  cell{1}{2} = {r=2}{},
  cell{1}{3} = {c=3}{0.425\linewidth},
  cell{1}{6} = {c=3}{0.431\linewidth},
  cell{3}{1} = {r=3}{},
  cell{6}{1} = {r=3}{},
  cell{9}{1} = {r=3}{},
  hline{1,12} = {-}{0.08em},
  hline{2} = {3-8}{0.03em},
  hline{3,6,9} = {-}{0.05em},
}
$k$ & $L$ & Test Acc. @ Epoch of Max. Train Acc. &                    &                     & Max Test Acc. @ Epoch  &                   &                             \\
  &   & GAT$_S$                                & GAT                & GATE                & GAT$_S$                                 & GAT               & GATE                        \\
1 & 1 & 92.0@2082*                             & 91.2@6830*         & \textbf{93.2@3712*} & 93.2@1421                               & 92.0@9564         & \textbf{\underline{93.6@3511}} \\
  & 2 & 89.6@8524*                             & 88.0@8935          & \textbf{91.2@942}   & 91.6@5188                               & 92.8@4198         & \textbf{\underline{95.6@111}}  \\
  & 3 & 86.4@9180*                             & 88.8@997           & \textbf{92.8@618}   & 91.2@6994                               & \underline{92.8@437} & \textbf{97.2@82}            \\
2 & 2 & 88.8@6736*                             & \textbf{89.6@3907} & 88.8@467            & 93.2@151                                & \textbf{93.2@95}  & \underline{92.0@105}           \\
  & 3 & 82.0@7612                              & 89.2@1950          & \textbf{91.6@370}   & 91.6@1108                               & 93.2@856          & \textbf{\underline{95.2@189}}  \\
  & 4 & 84.8@4898                              & 82.4@739           & \textbf{87.2@639}   & \underline{88.0@1744}                      & 88.4@423          & \textbf{90.4@447}           \\
3 & 3 & 80.8@8670                              & 80.4@737           & \textbf{85.2@391}   & \underline{86.4@1578}                      & 88.8@285          & \textbf{92.0@47}            \\
  & 4 & 78.0@3012                              & 80.4@767           & \textbf{89.6@480}   & 86.8@1762                               & 85.6@469          & \textbf{\underline{91.6@139}}  \\
  & 5 & 80.0@6611                              & 74.4@1701          & \textbf{86.0@447}   & 85.6@921                                & 83.6@1098         & \textbf{\underline{91.2@243}}  
\end{tblr}
\end{table*}

\begin{figure*} [t]
\centering
\includegraphics[width=.75\linewidth]{Figures/alphaDistribution/alpha_ii_distribution_legend.png}\\
\begin{subfigure}{.5\textwidth}
  \centering
\frame{\includegraphics[width=.99\linewidth]{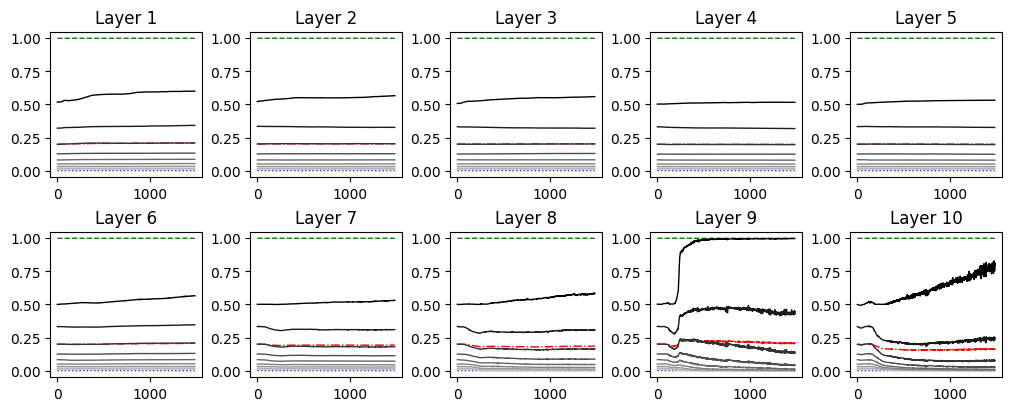}}
\caption{Tolokers, GAT: $61.6\%$ test AUROC.}
\end{subfigure}%
\begin{subfigure}{.5\textwidth}
  \centering  
\frame{\includegraphics[width=.99\linewidth]{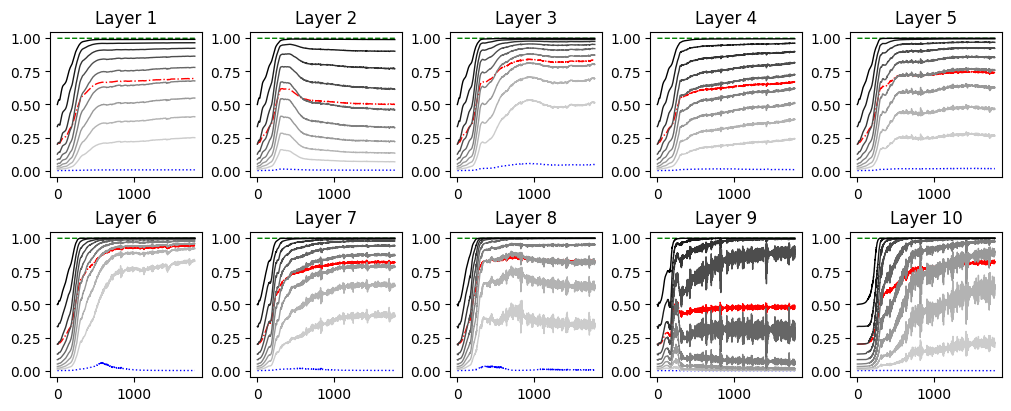}}
\caption{Tolokers, GATE: $69.2\%$ test AUROC.}
\end{subfigure}\\
  \caption{Distribution of $\alpha_{vv}$ against training epoch for one run of 10 layer networks on real-world heterophilic task. }
  \label{alpha-dist-gate-tolokers}
\end{figure*}

\begin{figure} [h]
\centering
\includegraphics[width=.75\linewidth]{Figures/alphaDistribution/alpha_ii_distribution_legend.png}\
\begin{subfigure}{.275\textwidth}
  \centering
\frame{\includegraphics[width=.99\linewidth]{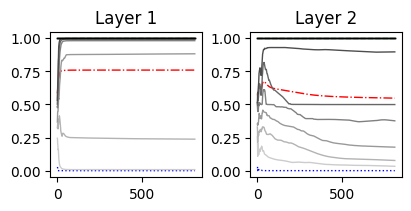}}
\end{subfigure}%
\begin{subfigure}{.675\textwidth}
  \centering  
\frame{\includegraphics[width=.99\linewidth]{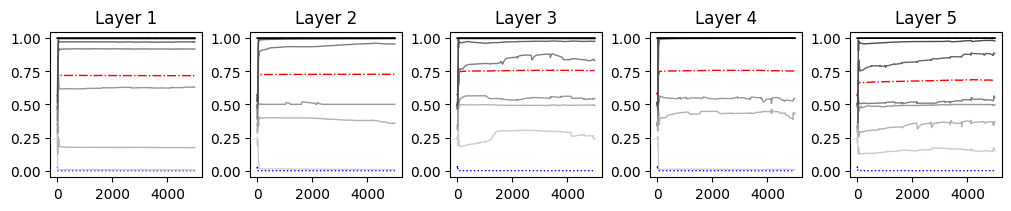}}
\end{subfigure}\\
\begin{subfigure}{.275\textwidth}
  \centering
\frame{\includegraphics[width=.99\linewidth]{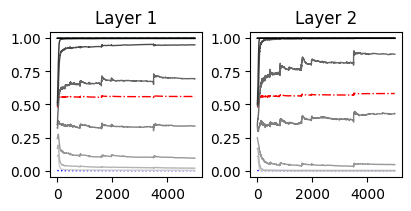}}
\end{subfigure}%
\begin{subfigure}{.675\textwidth}
  \centering  
\frame{\includegraphics[width=.99\linewidth]{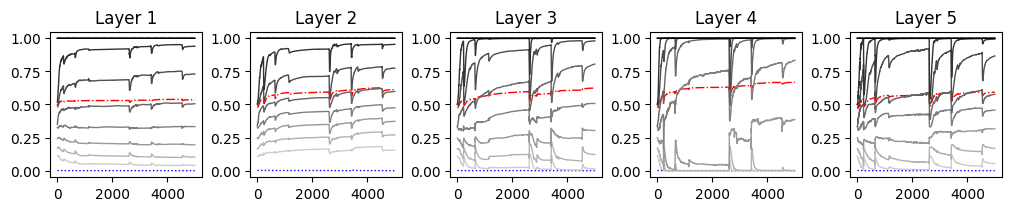}}
\end{subfigure}\\
  \caption{ Distribution of $\alpha_{vv}$ against training epoch of 2-layer (left) and 5-layer (right) GAT networks for heterophilic datasets Texas (top) and Actor (bottom) 2-layer modes. Despite having connections to unrelated neighbors, GAT is unable to switch off neighborhood aggregation.}
  \label{alpha-dist-gat-real-heterophilic}
\end{figure}

\begin{figure} [h]
\centering
\includegraphics[width=.75\linewidth]{Figures/alphaDistribution/alpha_ii_distribution_legend.png}\\
\begin{subfigure}{.275\textwidth}
  \centering
\frame{\includegraphics[width=.99\linewidth]{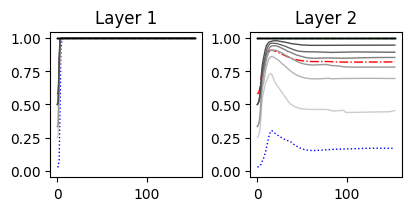}}
\end{subfigure}%
\begin{subfigure}{.675\textwidth}
  \centering  
\frame{\includegraphics[width=.99\linewidth]{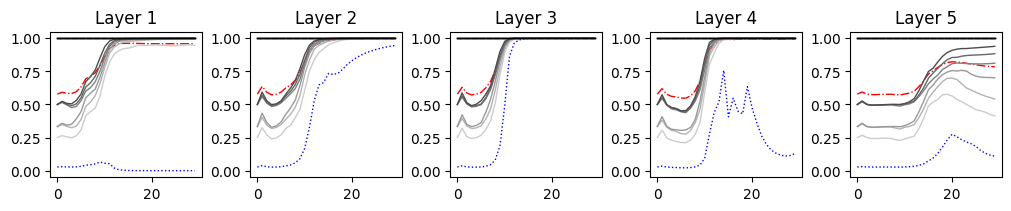}}
\end{subfigure}\\
\begin{subfigure}{.275\textwidth}
  \centering
\frame{\includegraphics[width=.99\linewidth]{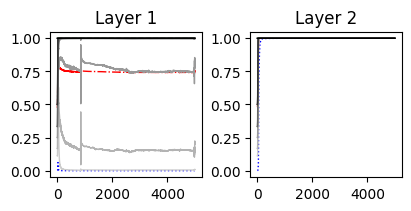}}
\end{subfigure}%
\begin{subfigure}{.675\textwidth}
  \centering  
\frame{\includegraphics[width=.99\linewidth]{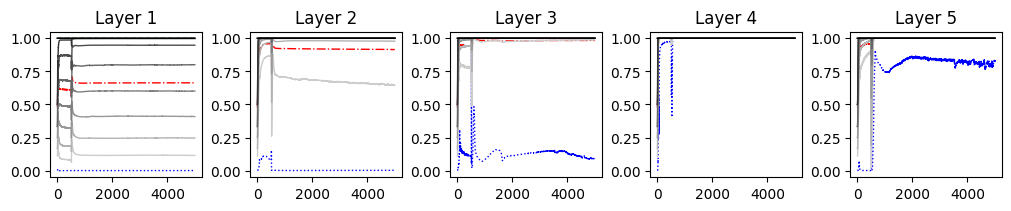}}
\end{subfigure}\\
  \caption{Distribution of $\alpha_{vv}$, against training epoch of 2-layer (left) and 5-layer (right) GATE networks for heterophilic datasets Texas (top) and Actor (bottom), across layers could be interpreted to indicate the inherent importance of raw node features relative to their neighborhoods. For instance, in the case of Texas, GATE carries out little to no neighborhood aggregation in the first layer over input node features. Instead, aggregation is mainly done over node features transformed in earlier layers that effectuate non-linear feature learning as in perceptrons. However, in the case of Actor, GATE prefers most of the neighborhood aggregation to occur over the input node features, indicating that they are more informative for the task at hand. }
  \label{alpha-dist-gate-real-heterophilic}
\end{figure}

\clearpage
\paragraph{Over-smoothing analysis} 

%define MAD, cite MAD, report MAD
In the main paper, we have already established the superior performance of GATE, compared to GAT, on several tasks. Intuitively, this can partially be attributed to reduced over-smoothing as its root cause, unnecessary neighborhood aggregation, is alleviated.
Here, we verify this insight quantitatively.

A widely accepted measure of over-smoothing is the Dirichlet energy (DE) \citep{cai2020note}. However, \citet{wGAT} propose a modification of DE to measure GAT energy $E_{GAT}$, that we use to evaluate over-smoothing in our experiments (see Table \ref{smoothness_DE}). We note that the notion of `over-smoothing' is itself task-dependent. It is difficult to determine the optimal degree of smoothing for a task and the threshold that determines `over'-smoothing. This merits an in-depth analysis and curation of task-dependent smoothness measures that are not our focus. To show that GATE reduced over-smoothing relative to GAT, it suffices that a decrease in smoothing and an increase in accuracy occur simultaneously.

\begin{table}[h]
    \caption{The measures  $E_{input}$, $E_{GAT}$ and $E_{GATE}$ denote the smoothness of input node features, node features at the last layer $L$ of the trained GAT and GATE models, respectively. Two cases are considered: All node pairs and only adjacent node pairs to measure smoothing at the global graph and local node level. Higher values indicate less smoothing. Node representations learned by GATE achieve higher test accuracy on all these tasks, as reported in the main paper, and are simultaneously less smooth than GAT in most cases, indicating that GATE potentially alleviates over-smoothing in GATs. }
    \label{smoothness_DE}
    \centering
    \begin{subtable}[h]{\textwidth}
    \caption{Synthetic self-sufficient task: varying graph structure and label distribution, node features as one-hot encoding of labels, $L=5$.}
    \label{DE_syn_self_suff}
    \centering
    \begin{tblr}{
  width = \linewidth,
  colspec = {Q[215]Q[117]Q[117]Q[117]Q[117]Q[117]Q[117]},
  cells = {c},
  cell{1}{1} = {r=2}{},
  cell{1}{2} = {c=3}{0.351\linewidth},
  cell{1}{5} = {c=3}{0.351\linewidth},
  hline{1,7} = {-}{0.08em},
  hline{2} = {2-7}{0.03em},
  hline{3} = {-}{0.05em},
}
Experiment setting  & All node pairs & &  &  Adjacent node pairs & & \\
                       &  $E_{input}$  & $E_{GAT}$ & $E_{GATE}$ &  $E_{input}$    &  $E_{GAT}$   & $E_{GATE}$ \\
Cora, Original Labels  & $ 6.016\;\mathrm{e}+06 $ & $ 1.281\;\mathrm{e}+08 $ & $ \mathbf{2.037\;\mathrm{e}+09} $ & $ 2.006\;\mathrm{e}+03 $ & $ 7.903\;\mathrm{e}+03 $ & $ \mathbf{1.971\;\mathrm{e}+05} $ \\
Cora, Random Labels    & $ 6.283\;\mathrm{e}+06 $ & $ 3.472\;\mathrm{e}+09 $ & $ \mathbf{3.747\;\mathrm{e}+09} $ & $ 9.080\;\mathrm{e}+03 $ & $ 9.306\;\mathrm{e}+05 $ & $ \mathbf{3.966\;\mathrm{e}+06} $ \\
ER$(p=.01)$, $2$ Classes     & $ 4.994\;\mathrm{e}+05 $ & $ 2.701\;\mathrm{e}+06 $ & $ \mathbf{4.478\;\mathrm{e}+07} $ & $ 5.042\;\mathrm{e}+03 $ & $ 2.272\;\mathrm{e}+04 $ & $ \mathbf{3.229\;\mathrm{e}+05} $ \\
ER$(p=.01)$, $8$ Classes      & $ 8.745\;\mathrm{e}+05 $ & $ 3.350\;\mathrm{e}+07 $ & $ \mathbf{2.615\;\mathrm{e}+08} $ & $ 8.694\;\mathrm{e}+03 $ & $ 1.762\;\mathrm{e}+05 $ & $ \mathbf{1.960\;\mathrm{e}+06} $ 
\end{tblr}
    \end{subtable}\\ \vspace{10pt}    
    \begin{subtable}[h]{\textwidth}
    \caption{Synthetic neighbor-dependent task:  graph structure, node features, generator parameter $k$, and label distribution as in Section \ref{sec:experiments}. }
    \label{DE_syn_nbr_dep}
    \centering
    \begin{tblr}{
  width = \linewidth,
  colspec = {Q[215]Q[117]Q[117]Q[117]Q[117]Q[117]Q[117]},
  cells = {c},
  cell{1}{1} = {r=2}{},
  cell{1}{2} = {c=3}{0.351\linewidth},
  cell{1}{5} = {c=3}{0.351\linewidth},
  hline{1,6} = {-}{0.08em},
  hline{2} = {2-7}{0.03em},
  hline{3} = {-}{0.05em},
}
Experiment setting  & All node pairs & &  &  Adjacent node pairs & &  \\
                       &  $E_{input}$  & $E_{GAT}$ & $E_{GATE}$ &  $E_{input}$    &  $E_{GAT}$   & $E_{GATE}$ \\
$k=1, L=3 $             & $ 1.953\;\mathrm{e}+06 $ & $ \mathbf{5.306\;\mathrm{e}+07} $ & $ 5.095\;\mathrm{e}+07 $ & $ 1.957\;\mathrm{e}+05 $ & $ \mathbf{4.234\;\mathrm{e}+05} $ & $ 3.610\;\mathrm{e}+05 $ \\
$k=2, L=4$               & $ 1.975\;\mathrm{e}+06 $ & $ 1.193\;\mathrm{e}+07 $ & $ \mathbf{2.198\;\mathrm{e}+07} $ & $ 2.012\;\mathrm{e}+04 $ & $ 1.016\;\mathrm{e}+05 $ & $ \mathbf{1.939\;\mathrm{e}+05} $ \\
$k=3, L=5$               & $ 1.951\;\mathrm{e}+06 $ & $ 1.645\;\mathrm{e}+07 $ & $ \mathbf{1.053\;\mathrm{e}+08} $ & $ 1.966\;\mathrm{e}+04 $ & $ 1.408\;\mathrm{e}+05 $ & $ \mathbf{9.096\;\mathrm{e}+05} $
\end{tblr}
    \end{subtable}\\ \vspace{10pt}    
    \begin{subtable}[h]{\textwidth}\caption{Real-world tasks.}
    \label{DE_real}
    \centering
    \begin{tblr}{
  width = \linewidth,
  colspec = {Q[150]Q[40]Q[117]Q[117]Q[127]Q[117]Q[117]Q[117]},
  cells = {c},
  cell{1}{1} = {r=2}{},
  cell{1}{2} = {r=2}{},
  cell{1}{3} = {c=3}{0.345\linewidth},
  cell{1}{6} = {c=3}{0.339\linewidth},
  hline{1,12} = {-}{0.08em},
  hline{2} = {3-8}{0.03em},
  hline{3} = {-}{0.05em},
}
Dataset           & $L$  
&  All node pairs & &  &  Adjacent node pairs & &  \\
                     &  &  $E_{input}$  & $E_{GAT}$ & $E_{GATE}$ &  $E_{input}$    &  $E_{GAT}$   & $E_{GATE}$ \\
roman-empire & $5$      & $ 1.274\;\mathrm{e}+09 $ & $ 1.002\;\mathrm{e}+11  $ & $ \mathbf{7.491\;\mathrm{e}+11} $ & $ 7.878\;\mathrm{e}+04 $ & $ 2.441\;\mathrm{e}+06 $ & $ \mathbf{4.009\;\mathrm{e}+07} $ \\
amazon-ratings & $10$         & $ 3.844\;\mathrm{e}+08 $ & $ 1.187\;\mathrm{e}+10  $ & $ \mathbf{2.272\;\mathrm{e}+10} $ & $ 4.933\;\mathrm{e}+04 $ & $ 3.848\;\mathrm{e}+05 $ & $ \mathbf{7.430\;\mathrm{e}+05} $ \\
minesweeper & $5$             & $ 6.869\;\mathrm{e}+07 $ & $ 1.386\;\mathrm{e}+09  $ & $ \mathbf{2.531\;\mathrm{e}+10} $ & $ 2.628\;\mathrm{e}+04 $ & $ 1.946\;\mathrm{e}+05 $ & $ \mathbf{7.017\;\mathrm{e}+06} $ \\
tolokers & $10$               & $ 1.391\;\mathrm{e}+08 $ & $ 1.044\;\mathrm{e}+11  $ & $ \mathbf{1.042\;\mathrm{e}+11} $ & $ 3.423\;\mathrm{e}+05 $ & $ 1.249\;\mathrm{e}+08 $ & $ \mathbf{1.397\;\mathrm{e}+08} $ \\
cora & $10$               & $ 5.088\;\mathrm{e}+05 $ & $ 1.437\;\mathrm{e}+07  $ & $ \mathbf{2.783\;\mathrm{e}+08} $ & $ 6.490\;\mathrm{e}+02 $ & $ 5.959\;\mathrm{e}+02 $ & $ \mathbf{1.226\;\mathrm{e}+04} $ \\
citeseer & $10$           & $ 3.463\;\mathrm{e}+05 $ & $ 2.916\;\mathrm{e}+05 $ & $ \mathbf{1.126\;\mathrm{e}+07} $ & $ 2.360\;\mathrm{e}+02 $ & $ 2.426\;\mathrm{e}+00 $ & $ \mathbf{1.030\;\mathrm{e}+02} $ \\
texas & $10$               & $ 1.945\;\mathrm{e}+06 $ & $ 3.280\;\mathrm{e}+04  $ & $ \mathbf{3.877\;\mathrm{e}+04} $ & $ 1.758\;\mathrm{e}+04 $ & $ 8.487\;\mathrm{e}+01 $ & $ \mathbf{9.695\;\mathrm{e}+01} $ \\
actor & $10$               & $ 2.612\;\mathrm{e}+08 $ & $ \mathbf{1.800\;\mathrm{e}+07}  $ & $ 1.364\;\mathrm{e}+07 $ & $ 1.237\;\mathrm{e}+05 $ & $ \mathbf{2.810\;\mathrm{e}+03} $ & $ 2.215\;\mathrm{e}+03 $ \\
wisconsin & $10$           & $ 4.438\;\mathrm{e}+06 $ & $ 1.057\;\mathrm{e}+07  $ & $ \mathbf{1.008\;\mathrm{e}+08} $ & $ 3.299\;\mathrm{e}+04 $ & $ 3.765\;\mathrm{e}+04 $ & $ \mathbf{7.363\;\mathrm{e}+08} $ 
\end{tblr}
    \end{subtable}
\end{table}

\clearpage
\paragraph{Comparison with other GNNs}

Other GNN architectures could potentially switch off neighborhood aggregation, as we show here. However, they are less flexible in assigning different importance to neighbors, suffer from over-smoothing, or come at the cost of an increased parameter count by increasing the size of the hidden dimensions (e.g. via a concatenation operation).
We evaluate the performance of three such architectures that, in principle, employ different aggregation methods, which are likely to be capable of switching off neighborhood aggregation, on synthetic datasets empirically and discuss their ability or inability to switch off neighborhood aggregation qualitatively as follows.

\begin{figure} [h]
\centering
\includegraphics[width=.75\linewidth]{Figures/alphaDistribution/alpha_ii_distribution_legend.png}
\begin{subfigure}{.125\textwidth}
  \centering
\frame{\includegraphics[width=.97\linewidth]{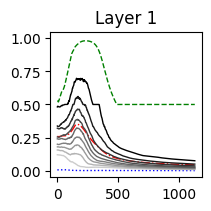}}
\end{subfigure}%
\begin{subfigure}{.25\textwidth}
  \centering  
\frame{\includegraphics[width=.95\linewidth]{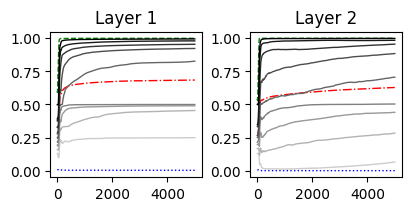}}
\end{subfigure}%
\begin{subfigure}{.625\textwidth}
  \centering  
\frame{\includegraphics[width=.95\linewidth]{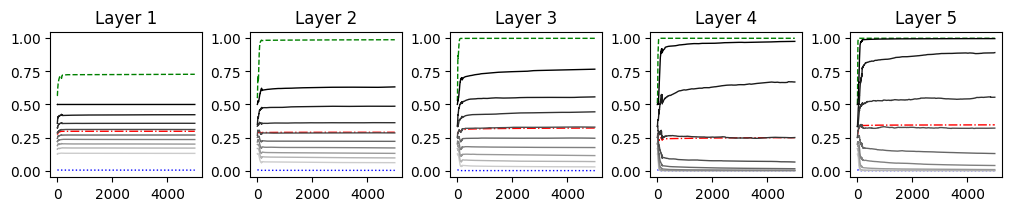}}
\end{subfigure}
  \caption{Distribution of $\alpha_{vv}$ against training epoch for self-sufficient learning problem using the Cora structure with random labels, where input node features are a one-hot encoding of node labels, for the $\omega$GAT architecture for the $1$, $2$ and $5$ layer models that achieve test accuracy of $100\%$, $98.5\%$, and $49.3\%$, respectively.}
  \label{nbr-aggr-wGAT}
\end{figure}

\begin{enumerate}
    \item $\omega$GAT \citep{wGAT} introduces an additional feature-wise layer parameter $\omega$ that can, in principle, switch off neighborhood aggregation by setting $\omega$ parameters to $0$, in addition to the attention mechanism based on GAT. However, in practice, as we verify on our synthetic dataset in Figure \ref{nbr-aggr-wGAT}, it is unable to effectively switch off neighborhood aggregation. Although it outperforms GAT, it is still substantially worse than GATE, especially for the deeper model due to unnecessary neighborhood aggregations. Another architecture based on graph attention, superGAT\cite{superGAT}, falls under the paradigm of structural learning as it uses a self-supervised attention mechanism essentially for link prediction between nodes, and therefore its comparison with GATE is infeasible.

\item GraphSAGE \citep{gcnHamilton} uses the concatenation operation to combine the node's own representation with the aggregated neighborhood representation. Therefore, it is usually (but not always) able to switch off the neighborhood aggregation for the synthetic datasets designed for the self-sufficient learning task (see Table \ref{self-suff-fagcn-sage}). Mostly, GATE performs better on the neighbor-dependent task, in particular for deeper models, where the performance of GraphSAGE drops likely due to over-smoothing (see Table \ref{nbr-dep-fagcn-sage}).

\item FAGCN \citep{fagcn} requires a slightly more detailed analysis. 
    Authors of FAGCN state in the paper that: `When $\alpha^{G}_{ij}\approx 0$, the contributions of neighbors will be limited, so the raw features will dominate the node representations.' where $\alpha^{G}_{ij}$ defined in the paper can be considered analogous to $\alpha_{ij}$ in GAT, though they are defined differently.
    Thus, from an expressivity point of view, FAGCN should be able to assign parameters such that all $\alpha^{G}_{ij}=0$.
    However, we empirically observe on synthetic datasets designed for the self-sufficient learning task, values of $\alpha^{G}_{ij}$ do not, in fact, approach zero. 
    Despite being unable to switch off neighborhood aggregation, FAGCN, in its default implementation, achieves $100\%$ test accuracy on the task. We discover this is so because FAGCN introduces direct skip connections of non-linearly transformed raw node features to every hidden layer. 
    Given the simplicity of the one-hot encoded features in the datasets and the complete dependence of the label on these features, FAGCN is able to represent the desired function. In order to better judge its ability to {\it switch off neighborhood aggregation by setting $\alpha^{G}_{ij}=0$}, we remove this skip connection. 
    From an expressivity point of view, FAGCN should still be able to achieve $100\%$ test accuracy by using only the (non-)linear transformations of raw features initially and performing no neighborhood aggregation in the hidden layers. 
    However, we find that FAGCN was unable to emulate this behavior in practice. 
    For a fair comparison of the differently designed attention mechanism in FAGCN with GATE, we introduce self-loops in the data so FAGCN may also receive a node's own features in every hidden layer. 
    Even then, FAGCN fails to achieve perfect test accuracy as shown in Table \ref{self-suff-fagcn-sage}. 
    Therefore, we suspect the attention mechanism in FAGCN may also be susceptible to the trainability issues we have identified for the attention mechanism in GAT. 
    Nevertheless, the capacity of FAGCN to learn negative associations with neighboring nodes is complementary to GATE and both could be combined. 
    It would be interesting to derive conservation laws inherent to other architectures such as FAGCN and GraphSAGE and study how they govern the behaviour of parameters.  
    Furthermore, by design, FAGCN does not perform any non-linear transformations of aggregated neighborhood features which may be necessary in some tasks, such as our synthetic dataset for the neighbor-dependent learning task. As Table \ref{nbr-dep-fagcn-sage} shows, GATE outperforms FAGCN on such a task.

\end{enumerate}

Lastly, we would like to emphasize that our aim is to provide insights into the attention mechanism of GAT and understand its limitations. 
While it should be able to flexibly assign importance to neighbors and the node itself without the need for concatenated representation or explicit skip connections of the raw features to every layer, it is currently unable to do so in practice. 
In order to verify our identification of trainability issues, we modify the GAT architecture to enable the trainability of attention parameters which control the trade-off between node features and structural information.

 \begin{table}[h]
\caption{Self-sufficient learning: $S, C$ and $L$ denote graph structure, number of label classes, and number of network layers, respectively. Original (Orig.) and Randomized (Rand.) labels are used for the Cora structure. The FAGCN model is implemented without skip connections from the input layer to every other layer and without any self-loops in input data, whereas FAGCN* denotes the model also without skip connections but with self-loops introduced for all nodes in input data.}
\label{self-suff-fagcn-sage}
\centering
\begin{tblr}{
  width = \linewidth,
  colspec = {Q[138]Q[54]Q[90]Q[127]Q[127]Q[127]Q[127]Q[127]},
  cells = {c},
  cell{1}{1} = {r=2}{},
  cell{1}{2} = {r=2}{},
  cell{1}{3} = {r=2}{},
  cell{1}{4} = {c=5}{0.635\linewidth},
  cell{3}{1} = {r=6}{},
  cell{3}{2} = {r=3}{},
  cell{6}{2} = {r=3}{},
  cell{9}{1} = {r=6}{},
  cell{9}{2} = {r=3}{},
  cell{12}{2} = {r=3}{},
  hline{1,15} = {-}{0.08em},
  hline{2} = {4-8}{0.03em},
  hline{3,9} = {-}{0.05em},
  hline{6,12} = {2-8}{0.03em},
}
Structure & $C$ & $L$ & Max. Test Accuracy ($\%$) &  &  &  & \\
 &  &  & GAT & GATE & FAGCN & FAGCN* & SAGE\\
Cora & O,7 & $1$ & $\mathbf{100}$ & $\mathbf{100}$ & $90.1$ & $97.6$ & $\mathbf{100}$\\
 &  & $2$ & $94.6$ & $\mathbf{100}$ & $94.2$ & $94.9$ & $98.8$\\
 &  & $5$ & $88.5$ & $\mathbf{99.7}$ & $87.1$ & $89.1$ & $92.4$\\
 & R,7 & $1$ & $\mathbf{100}$ & $\mathbf{100}$ & $61.6$ & $97.8$ & $\mathbf{100}$\\
 &  & $2$ & $57.0$ & $\mathbf{100}$ & $69.2$ & $70.5$ & $\mathbf{100}$\\
 &  & $5$ & $36.7$ & $\mathbf{100}$ & $21.2$ & $36.7$ & $99.6$\\
ER ($p=0.01$) & R,2 & $1$ & $\mathbf{100}$ & $\mathbf{100}$ & $\mathbf{100}$ & $\mathbf{100}$ & $\mathbf{100}$\\
 &  & $2$ & $\mathbf{100}$ & $\mathbf{100}$ & $\mathbf{100}$ & $\mathbf{100}$ & $\mathbf{100}$\\
 &  & $5$ & $99.6$ & $\mathbf{100}$ & $96.4$ & $99.2$ & $\mathbf{100}$\\
 & R,8 & $1$ & $99.2$ & $\mathbf{100}$ & $86.4$ & $98.8$ & $\mathbf{100}$\\
 &  & $2$ & $97.6$ & $\mathbf{100}$ & $86.0$ & $91.6$ & $\mathbf{100}$\\
 &  & $5$ & $38.4$ & $\mathbf{100}$ & $31.6$ & $40.4$ & $\mathbf{100}$
\end{tblr}
\end{table}

 \begin{table}[h]
\caption{Neighbor-dependent learning: $k$ and $L$ denote the number of hops aggregated in the neighborhood to generate labels, and the number of layers of the evaluated network, respectively.}
\label{nbr-dep-fagcn-sage}
\centering
\begin{tblr}{
  width = \linewidth,
  colspec = {Q[77]Q[77]Q[169]Q[169]Q[169]Q[169]},
  cells = {c},
  cell{1}{1} = {r=2}{},
  cell{1}{2} = {r=2}{},
  cell{1}{3} = {c=4}{0.676\linewidth},
  cell{3}{1} = {r=3}{},
  cell{6}{1} = {r=3}{},
  cell{9}{1} = {r=3}{},
  hline{1,12} = {-}{0.08em},
  hline{2} = {3-6}{0.03em},
  hline{3,6,9} = {-}{0.05em},
}
$k$ & $L$ & Max Test Accuracy ($\%$) $@$ Epoch  &  &  & \\
 &  & GAT & GATE & SAGE & FAGCN\\
$1$ & $1$ & $92 @ 9564$ & $\textbf{93.6 @ 3511}$ & $93.2 @ 2370$ & $93.2 @ 1618$\\
 & $2$ & $92.8 @ 4198$ & $\textbf{95.6 @ 111}$ & $95.6 @ 723$ & $94.1 @ 1455$\\
 & $3$ & $92.8 @ 437$ & $\textbf{97.2 @ 82}$ & $96.8 @ 100$ & $81.2 @ 573$\\
$2$ & $2$ & $\textbf{93.2 @ 95}$ & $92.0 @ 105$ & $90.8 @ 199$ & $90.4 @ 170$\\
 & $3$ & $93.2 @ 856$ & $\textbf{95.2 @ 189}$ & $94.4 @ 113$ & $88.8 @ 283$\\
 & $4$ & $88.4 @ 423$ & $90.4 @ 447$ & $\textbf{92.4 @ 139}$ & $87.6 @ 549$\\
$3$ & $3$ & $88.8 @ 285$ & $\textbf{92.0 @ 47}$ & $87.6 @ 45$ & $89.2 @ 528$\\
 & $4$ & $85.6 @ 469$ & $\textbf{91.6 @ 139}$ & $88 @ 60$ & $89.2 @ 3191$\\
 & $5$ & $83.6 @ 1098$ & $\textbf{91.2 @ 243}$ & $86.0 @ 35$ & $88.8 @ 205$
\end{tblr}
\end{table}

%%%%%%%%%%%%%%%%%%%%%%%%%%%%%%%%%%%%%%%%%%%%%%%%%%%%%%%%%%%%%%%%%%%%%%%%%%%%%%%
%%%%%%%%%%%%%%%%%%%%%%%%%%%%%%%%%%%%%%%%%%%%%%%%%%%%%%%%%%%%%%%%%%%%%%%%%%%%%%%

\end{document}